\documentclass{article}

\PassOptionsToPackage{numbers, compress}{natbib}

    \usepackage[preprint]{neurips_2025}

\usepackage[utf8]{inputenc} 
\usepackage[T1]{fontenc}    
\usepackage{hyperref}       
\usepackage{url}            
\usepackage{booktabs}       
\usepackage{amsfonts}       
\usepackage{nicefrac}       
\usepackage{microtype}      
\usepackage{xcolor}         
\usepackage{siunitx}
\usepackage{amsmath}        
\usepackage{bm} 
\usepackage{graphicx}
\usepackage{subcaption}
\usepackage{float}
\usepackage{multirow}
\usepackage{longtable}
\usepackage{doi}

\title{Friction on Demand: A Generative Framework for the Inverse Design of Metainterfaces}

\author{%
  Valentin~Mouton\\
  École centrale de Lyon\\
  LTDS, UMR 5513\\
  Écully, France \\
  \texttt{valentin.mouton@ec-lyon.fr} \\
  \And
  Adrien~Mélot \\
  Univ. Gustave Eiffel, Inria \\
  COSYS-SII, I4S \\
  Rennes, France \\
  \texttt{adrien.melot@inria.fr} \\
}

\begin{document}

\maketitle

\begin{abstract}

Designing frictional interfaces to exhibit prescribed macroscopic behavior is a challenging inverse problem, made difficult by the non-uniqueness of solutions and the computational cost of contact simulations. Traditional approaches rely on heuristic search over low-dimensional parameterizations, which limits their applicability to more complex or nonlinear friction laws. We introduce a generative modeling framework using Variational Autoencoders (VAEs) to infer surface topographies from target friction laws. Trained on a synthetic dataset composed of 200~million samples constructed from a parameterized contact mechanics model, the proposed method enables efficient, simulation-free generation of candidate topographies. We examine the potential and limitations of generative modeling for this inverse design task, focusing on balancing accuracy, throughput, and diversity in the generated solutions. Our results highlight trade-offs and outline practical considerations when balancing these objectives. This approach paves the way for near-real-time control of frictional behavior through tailored surface topographies.
\end{abstract}

\section{Introduction}

Designing interfaces with tailored frictional behavior is a longstanding challenge in contact mechanics, with strong implications for a wide range of technologies, including robotic manipulation, haptic devices, soft material systems, brakes, etc. This difficulty stems from the multiscale nature of the surface roughness, the complexity of interfacial contact mechanics, and the general lack of analytical tractability in tribological models. While recent work experimentally demonstrated the feasibility of designing metainterfaces~\cite{Aymard2024,Aymard2023Thesis}, determining an asperity distribution or surface topography that yields a target friction law is ill-posed, as the mapping from target friction laws to surface topography is often non-unique and the feasible design space is discontinuous. Early approaches to friction tuning relied on either manual exploration of the design space~\cite{murarash2011tuning, li2016tuning} or the analytical inversion of simplified models~\cite{Aymard2024,Aymard2023Thesis}, which required assuming low-dimensional topographic parameterizations. Thus, these approaches do not extend naturally to more complex design tasks, such as those involving more complex and flexible friction laws or higher-dimensional surface representations.

Conventional approaches to this problem rely on heuristic search algorithms to navigate the design space. These methods require repeated evaluations of computationally expansive contact models to assess each candidate design, which severely limits their applicability in near-real-time or high-throughput settings. Furthermore, heuristic methods struggle to capture the
multimodal and high-dimensional nature of the design space, often converging to suboptimal solutions or failing to explore the full range of feasible designs. As a result, the design of frictional interfaces remains heavily reliant on empirical tuning and trial-and-error, even in applications where precise control of frictional behavior is essential for system performance and reliability.

In this work, we move beyond this limitation by proposing a data-driven inference framework. We introduce a Conditional Variational Autoencoder (CVAE) trained on a large-scale, 200-million-sample synthetic dataset to directly learn the inverse mapping from target friction laws to their corresponding surface topographies. To the best of our knowledge, this dataset is among the largest for scientific machine learning and the second-largest synthetic tabular dataset by sample size, after ClimSim~\cite{yu2023climsim}. We provide a comprehensive analysis of this framework and examine the trade-offs between prediction accuracy, solution diversity, and inference speed. Our results highlight both the potential of generative modeling for metainterface design and the limitations concerning absolute functional fidelity, outlining a path toward practical, near-real-time control of friction through tailored interface topography.

\section{Related Works}

While machine learning models have been applied across many domains, including tribology~\cite{zhu2022generative,kaliafetis2025using}, their use for designing metainterfaces with tailored friction laws remains unexplored. This section situates our contributions within the generative inverse design literature and highlights four often overlooked challenges: \textbf{(i)} reassessing the role of so-called outdated architectures, which are often disregarded simply because they are no longer state-of-the-art in some domains, yet may be better suited to scientific regression problems; \textbf{(ii)} handling data distributions that fall outside conventional machine learning standards; \textbf{(iii)} bridging the gap between parameter-level and functional-level accuracy; \textbf{(iv)} ensuring generalization from synthetic training data to real-world experimental targets.

\textbf{(i) Generative Models for Inverse Design}. Generative models were first popularized in image generation, where VAEs and CVAEs provided some of the earliest frameworks for learning latent representations. However, their use in image generation has declined because they tend to average pixels and produce blurry reconstructions that lack high-frequency details. They were soon surpassed by Generative Adversarial Networks (GANs), with architectures such as StyleGANs achieving state-of-the-art photorealism~\cite{karras_stylegan3}, or by diffusion models, which have emerged as the dominant approach for image generation, because they support direct inference from text prompting, which GANs cannot easily provide~\cite{sauer_stylegant_2023}. Such generative models have since been readily adapted to scientific inverse design tasks across domains such as photonics~\cite{luiphotonicmetasurface,acsphotonics2025diffusion}, molecular design~\cite{sanchez2018inverse} and mechanical engineering~\cite{bastek2023inverse,brzin2024}. However, most studies and ML benchmarks treat the problem as an image-to-image or image-to-parameter task, where generative models benefit from strong local correlations, since neighboring pixels in an image carry meaningful information that can be compressed into low-dimensional latent spaces. \textit{This has led to a strong bias in the field, since newer, more complex architectures, such as Normalizing Flows and Diffusion Models are widely assumed to be inherently superior, with their success in image generation benchmarks often taken as an absolute truth that carries over to all other tasks.} However, not all inverse design problems are created equal. In our case, we tackle an inverse regression problem and the mapping between surface topographies and friction laws exhibits no obvious local correlations: two surfaces that yield the same friction law may be completely uncorrelated in parameter space, and conversely, small changes in surface topography can lead to drastically different frictional behavior. This absence of local structure makes our problem significantly harder for generative models than image-based problems. Our work provides a in-depth analysis of generative models in inverse regression tasks for tribological applications. 

\textbf{(ii) Challenges in Modeling Scientific Data Distributions.}~In many scientific domains, \textit{data distributions are shaped by the way they are generated rather than by the central limit theorem} as in computer vision and NLP. In our case, exhaustive coverage of the design space through parameter sweeps produces a uniformly sampled input space, which poses a challenge for generative models that are architecturally biased. For instance, VAEs adopt Gaussian priors for mathematical convenience, as this enables training with a closed-form Kullback–Leibler (KL) divergence regularization term~\cite{higgins2017beta-kl}. Alternatives such as VampPrior~\cite{tomczak2018vaevampprior}, Diffusion Models with learnable priors~\cite{zhao2024ccdpm}, or specialized architectures like UniGAN~\cite{pan2022unigan} have been proposed, but they often introduce a significant computational overhead and training complexity. Our work departs from this trend: we train on a uniformly distributed 200-million-sample dataset, to our knowledge the second-largest synthetic tabular dataset by sample size after ClimSim~\cite{yu2023climsim}, and we demonstrate that simply enlarging the latent dimension of a standard VAE or CVAE suffices to achieve near-perfect parameter reconstruction under uniform distributions. This suggests that, when both data scale and latent capacity are sufficient, complex priors or models are not always necessary.

\textbf{(iii) From Parameter-Level to Functional-Level Accuracy}.~A major challenge in applying these models to scientific problems is ensuring that high accuracy on benchmark metrics translates to real-world utility. A model can reconstruct a system's parameters with high precision yet fail to reproduce its associated functional behavior. This disconnect is recognized in some fields, e.g. CPU simulator tuning~\cite{Renda2020DiffTune}, protein design~\cite{huang2025eva}, hardware acceleration~\cite{huang2022learning} and inverse optimization theory~\cite{ren2025inverse} and several works have attempted to optimizing for functional scores, such as "synthetic complexity," rather than just parameter reconstruction~\cite{Coley2018SCScore, Thakkar2021RAscore}. However, many studies stop at good-enough representations without assessing the end-to-end functional accuracy~\cite{brzin2024,acsphotonics2025diffusion}.Our work contributes to this area by investigating the extent to which reaching a high accuracy in learning an intermediate representation based on a Gaussian Mixture Model (GMM) translates to the accuracy of the final functional to be evaluated.

\textbf{(iv) Generalization and Sim-to-Real Gap.}~In ML, generalization usually refers to a model’s ability to perform well on unseen data drawn from a similar distribution as the training set (often a reserved part of the dataset). Standard evaluation protocols for inverse design solvers follow this convention: they train and test on synthetic or real datasets sampled from a single distribution, with success measured on a held-out test set with known ground truth~\cite{daras2024warped, khaireh2023newcomer}. While this validates in-distribution generalization, it does not ensure robustness to out-of-distribution (OOD) inputs that differ from the training data. For instance, in computer vision and image generation, a generative model trained on a class, e.g. human faces, will not be able to generate another, e.g. dogs, since the latter fall outside the training distribution~\cite{karras_stylegan3,sauer_styleganxl}. In our case, after training on a 200-million-sample synthetic dataset derived from a parameterized tribological model, we perform a zero-shot functional test on a target friction law derived from physical experiments whose interface topography lies outside of the parameter bounds of the dataset. Unlike the vision analogy, however, this does not involve switching between different data classes, as all our samples correspond to friction laws. The OOD challenge here arises from the need to extrapolate beyond the support of the training distribution, while the underlying problem remains unchanged. This OOD test directly measures whether the learned representation transfers to real systems, a criterion which is rarely addressed in the inverse design literature.

\section{Method}

\textbf{Main Assumptions.}~The design of frictional metainterfaces requires modeling the relationship between surface topography and macroscopic friction law. Without loss of generality, we adopt herein the same assumptions as those
used in a recent experimental proof of concept~\cite{Aymard2024, Aymard2023Thesis}, which successfully designed glass–elastomer interfaces with prescribed friction laws. This choice ensures comparability with a validated experimental framework, which should facilitate the translation of our generative modeling framework to practical tribological applications while maintaining generality across non-adhesive elastic contact systems.
Specifically, our approach is inspired by the Greenwood and Williamson (GW) model~\cite{Greenwood1966}, which treats the contact between two rough surfaces as the contact between an equivalent rough surface and a rigid, flat plane. As the GW model assumes non-adhesive elastic contacts, it is well-suited for glass–elastomer interfaces. We extend this framework by employing a more descriptive surface topography model and by directly simulating the resulting forces, in order to exhaustively explore the design space.

\textbf{Problem Description.}~A central challenge in the inverse design of frictional metainterfaces is the accurate representation of surface topography. Since describing each individual asperity is computationally intractable, statistical models are employed. Unlike the classic GW model, which assumes simple, independent distributions for asperity properties (i.e., an exponential height distribution and a constant radius of curvature),  we model the joint distribution of asperity height \(h\) and radius of curvature \(R\) using a novel approach based on Gaussian Mixture Models (GMM).  The GMM is parameterized by \(\bm{\theta}\), which contains the mixture weights, means, and covariances of the \mbox{Gaussian} components. This approach allows for the representation of complex, multi-modal surface topographies~\cite{Silva2024}.  

\textbf{Problem Formulation.}~The inverse problem can then be formulated as follows: given a desired friction law \(\bm{F}(P)\), the task is to identify the GMM parameters \(\bm{\theta}\) that generate a surface topography consistent with this target law.   

\textbf{Dataset.}~We constructed a 200-million-sample dataset by sampling the GMM parameter space \(\bm{\theta}\) and computing the corresponding friction laws \(\bm{F}(P)\) through forward simulations of asperity-level contact mechanics. Each sample consists of a 23-dimensional parameter vector \(\bm{\theta}\), an asperity count \(N\), and the discretized friction law generated from the resulting surface realization. To ensure a quasi-uniform coverage of the design space, the GMM parameters are sampled using Sobol sequences~\cite{sobol1967distribution} within prescribed bounds, resulting into quasi-linearly independent variables (see Appendix~\ref{appendix:dataset}, Figure~\ref{fig:corr_yy}). For each \(\bm{\theta}\), discrete asperities are drawn from the corresponding GMM and the resulting friction forces are computed over a range of normal forces to obtain the friction law. Full details of the computational implementation are provided in Appendix~\ref{appendix:dataset}.

\textbf{Model Architecture.}~The mapping from a target friction law \(\bm{F}(P)\) to feasible GMM parameters \(\bm{\theta}\) is ill-posed as different surface topographies may yield similar friction laws. A purely deterministic regression would collapse this diversity into a single estimate, discarding valid solutions. To address this, we employ a generative approach capable of representing conditional distributions over \(\bm{\theta}\). Variational Autoencoders~\cite{kingma2013vae} are particularly well suited to this task, as they learn a probabilistic latent representation of the solution space and enable sampling of multiple candidates consistent with the same input. By conditioning a VAE on \(\bm{F}(P)\), the resulting CVAE allows for generating GMM parameters given a target friction law. Although the latent prior in VAEs is Gaussian, we handle the quasi-uniform distribution of the GMM parameters by expanding the latent dimension, as discussed later in the paper.

\textbf{Training Objective.}~The CVAE is trained to minimize the sum of a reconstruction loss and a weighted Kullback--Leibler (KL) divergence term, following the standard \(\beta\)-VAE approach~\cite{higgins2017beta-kl}. The reconstruction loss is computed using a Smooth L1 (Huber) function~\cite{huber1964robust} between the original and reconstructed GMM parameters, while the KL term regularizes the latent space to align with a Gaussian prior. KL annealing is employed by gradually increasing the weight \(\beta_{KL}\) from near zero, which improves training stability and mitigates posterior collapse.

\textbf{Hyperparameter Optimization.}~We performed hyperparameter tuning using the Optuna framework~\cite{Akiba2019Optuna}, optimizing batch size, learning rate, weight decay, latent dimension, and network architecture. A total of 331 trials were completed, corresponding to 6,58 million steps. The total optimization time was approximately 85 hours (about 3.5 days). All experiments were conducted on a single Nvidia GeForce RTX 4060 Ti 16 GB GPU. Trials utilized a Tree-structured Parzen Estimator (TPE) sampler with median pruning. Detailed configurations are provided in~\ref{appendix:model} (Table~\ref{tab:cvae-optuna-methodology-full}).

\section{Results}

\subsection{CVAE Performance on GMM Parameters Prediction}

\textbf{Prediction Accuracy.} The primary evaluation concerns the CVAE’s ability to predict the GMM parameters from a target friction law. As shown in Table~\ref{tab:eval_results}, on a test set of randomly selected 30,031,872 samples (15\% of the dataset), the CVAE achieves a median Symmetric Mean Absolute Percentage Error (sMAPE) of only 2.27\% between the predicted and ground-truth GMM parameters. This strong performance is further corroborated by a uniform averaged adjusted \(R^2\) score of 0.9987, indicating that the model explains nearly all of the variance in the target parameters. The errors are concentrated, with 95\% of samples below 7.17\% sMAPE.

\begin{table}[htb]
  \centering
  \caption{Evaluation results on the full 30-million-sample test set}
  \label{tab:eval_results}
  \begin{tabular}{ccccc c c}
    \toprule
    \multicolumn{5}{c}{\textbf{Relative Error (sMAPE)}} & \textbf{Pred. Acc.} & \textbf{Distr. Similarity} \\
    \cmidrule(r){1-5} 
    P25 & Median & Mean & P75 & P99 & Adjusted \(R^2\) & Avg. Wasserstein Dist. \\
    \midrule
    1.687\% & 2.270\% & 2.947\% & 3.409\% & 11.27\% & 0.9987 & 0.0086 \\
    \bottomrule
  \end{tabular}
\end{table}

\textbf{Physical Validity.} Some of the model's raw outputs may not satisfy all physical constraints (e.g., mixture weights summing to one). In such cases, a clamping and normalization procedure (see Appendix~\ref{appendix:clamping}) enforces these constraints which ensures that all outputs are physically valid.  The average Wasserstein distance of 0.0086 between generated and test-set parameters (scaled space) suggests satisfactory alignment with the target distribution, although the \texttt{tanh} output activation introduces a slight U-shaped bias (Appendix~\ref{appendix:distrib}, Figure~\ref{fig:dist-coverage}). These results suggest that the CVAE has successfully learned a high-fidelity mapping from the functional domain to the parameter space.

\begin{figure}[h!]
    \centering
    \includegraphics[width=\linewidth]{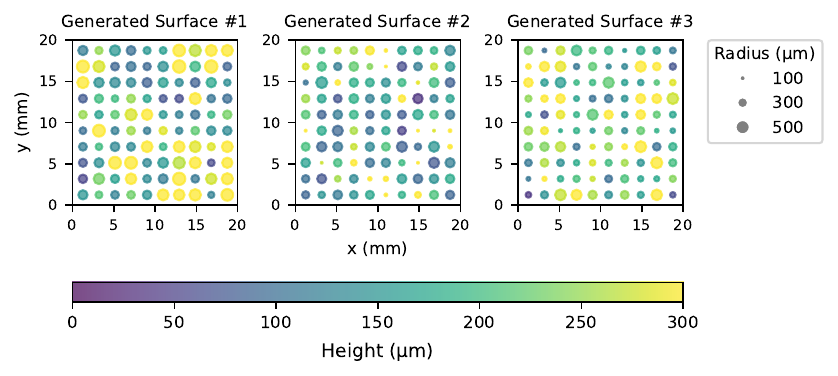}
    \caption{Representative surface topographies generated by the CVAE from three distinct latent samples for the same target, demonstrating the model's ability to capture the multimodality of the solution space.}
    \label{fig:topo}
\end{figure}

\textbf{Uncertainty and Diversity in CVAE Predictions.} The variational formulation enables sampling multiple parameter sets for the same target law. Figure~\ref{fig:cvae-uncertainty} (Appendix~\ref{appendix:uncertainty}) illustrates the mean predicted law with ±1~standard deviation from repeated latent space sampling, where the shaded region quantifies predictive uncertainty. Figure~\ref{fig:topo} presents three physically valid surface topographies generated for the same target. Despite differences in asperity distributions, all are predicted to yield similar macroscopic behavior, reflecting the multimodal nature of the inverse problem. We further investigated convergence with respect to the number of latent samples. The functional sMAPE stabilizes after 10,000 inferences, indicating that repeated sampling beyond this point provides little additional reduction in prediction error (see Appendix~\ref{appendix:convergence}, Figure~\ref{fig:cvae-convergence}).

\textbf{Ablations and Baseline Comparisons.} The selection of the CVAE and its final configuration was guided by a comprehensive set of experiments, including benchmarking against alternative architectures (see Table~\ref{tab:baseline_comparison}) and extensive ablation studies (detailed in Appendix~\ref{appendix:ablation}). We compared the CVAE to standard regression models (MLP and XGBoost~\cite{chen2016xgboost}). These baselines confirm that a generative approach is essential for accurately mapping the uniform distribution of GMM parameters, whereas deterministic models fail to do so, as reflected by adjusted \(R^2\) scores near zero of both MLP and XGBoost. A conditional GAN hyperparameter optimization was also conducted, but severe mode collapse prevented it from learning the quasi-uniform distributions, so its results are omitted. In contrast, the CVAE framework provided superior stability and predictive accuracy. Furthermore, our ablation studies investigated the impact of latent dimensionality, KL regularization, and other hyperparameters. They revealed that model performance was most sensitive to two factors: the latent dimension and whether conditioning inputs were used. This finding is particularly critical, as removing the conditioning, which reduces the model to a standard VAE, led to a significant improvement in parameter reconstruction accuracy and forms the basis of the high-fidelity benchmark analyzed in subsequent sections. The hyperparameters of the selected CVAE are listed in Appendix~\ref{appendix:model}.

\begin{table}[H]
\centering
\caption{Baseline comparison on the full 30-million samples test set.}
\label{tab:baseline_comparison}
\begin{tabular}{lccccc}
\toprule
Model & sMAPE (\%) & Adjusted \(R^2\) & Hardware & Inference Time\\
\midrule
VAE & 1.697 & 0.9997 & 1\,\(\times\)\,RTX 4060 Ti & 52.17s\\
CVAE & 2.968 & 0.9987 & 1\,\(\times\)\,RTX 4060 Ti & 51.42s\\
XGBoost & 63.97 & 0.0454 & 1\,\(\times\)\,RTX 4060 Ti & 12h 53m\\
MLP & 67.68 & -0.0129 & 1\,\(\times\)\,RTX 4060 Ti & 41.16s\\
\bottomrule
\end{tabular}
\end{table}

\subsection{From Parameter Accuracy to Functional Error: A Performance Discrepancy} 

While the CVAE achieves low parameter-level error, the ultimate criterion is fidelity of the friction law. Because the forward simulation from surface parameters to friction response is nonlinear, even small parameter deviations can lead to large functional errors. We therefore assess end-to-end functional error and contrast amortized inference (CVAE) with optimization-based inference via the VAE obtained from the ablation study coupled with the CMA-ES optimizer~\cite{hansen2006cma}.

\begin{figure}[h!] 
    \centering
    
    \begin{subfigure}[b]{0.45\linewidth}
        \centering
        \includegraphics[width=\linewidth]{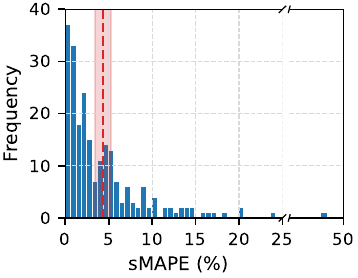}
        \caption{VAE\,+\,CMA-ES distribution.}
        \label{fig:dist-vae}
    \end{subfigure}
    \hfill 
    \begin{subfigure}[b]{0.45\linewidth}
        \centering
        \includegraphics[width=\linewidth]{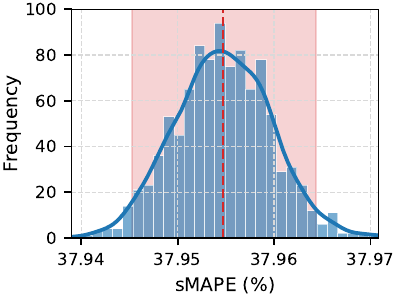}
        \caption{CVAE distribution.}
        \label{fig:dist-cvae}
    \end{subfigure}
    
    \caption{
        Comparison of sMAPE error distributions. (\subref{fig:dist-vae}) Distribution for the VAE\,+\,CMA\mbox{-}ES method over 225 randomly selected test samples. Mean error and 99\% confidence interval are shown by the dashed red line and shaded region. The x-axis is broken to show the full range of data. (\subref{fig:dist-cvae}) Error distribution for the CVAE. Mean error and 95\% confidence interval, estimated via 1,000 bootstrap resamples, are shown by the dashed red line and shaded region.
    }
    \label{fig:VAE_CVAE_sMAPE_comparison}
\end{figure}

\textbf{Comparison on Functional sMAPE.}~Figures~\ref{fig:dist-vae} and~\ref{fig:dist-cvae} show the functional sMAPE distributions for VAE\,+\,CMA\mbox{-}ES and CVAE, respectively, computed on the predicted friction laws. For~VAE\,+\,CMA\mbox{-}ES, the distribution is right-skewed, with a median of 2.50\% and a mean of 4.35\%, while the maximum error reaches 48.11\%, indicating occasional large deviations. Confidence interval analysis estimates the true mean sMAPE over the full 30-million-sample population to lie between 3.41\% and 5.29\% (99\% CI, analytical estimate). Performing inference on a randomly selected set of 225 friction laws from the test set required approximately 45~minutes, using 75 CMA-ES iterations per law and initializing near the true GMM parameters to accelerate convergence. In contrast, the CVAE exhibits a higher average error (37.95\%) but an extremely narrow distribution, with low variance estimated from 1,000 bootstrap resamples of the full 30-million-row test set (\(95\% \ \text{CI} \ [37.945\%, 37.965\%]\)), indicating a higher bias. End-to-end inference with the CVAE, implemented in JAX (see Appendix~\ref{appendix:dataset}), over the full test set required approximately 3\text{~h}~15\text{~min}. These results highlight a clear trade-off: VAE\,+\,CMA\mbox{-}ES achieves lower functional error at the cost of iterative optimization and higher variance, whereas the CVAE delivers fast, consistent predictions with a predictable bias.

\begin{figure}[h!]
    \centering
    \includegraphics[width=0.85\linewidth]{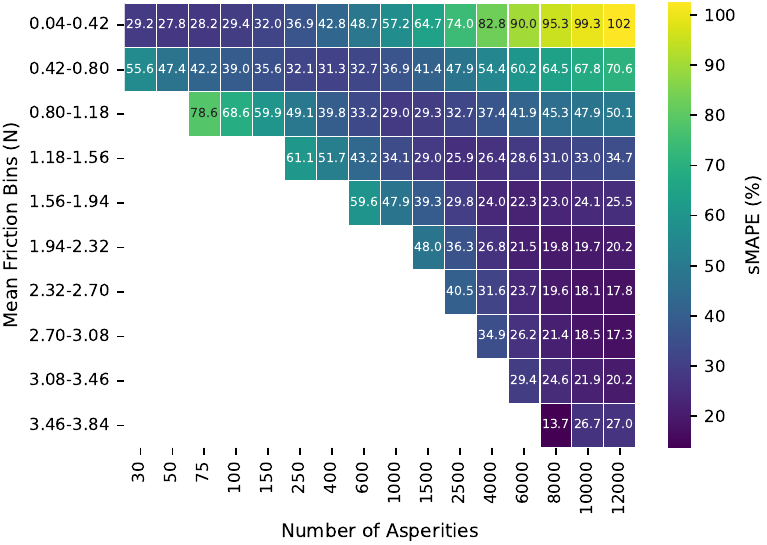}
    \caption{CVAE sMAPE performance across different physical regimes, defined by the number of asperities (x-axis) and the mean friction force (y-axis).}
    \label{fig:performance-heatmap}
\end{figure}

\textbf{Performance Across the Design Space.}~The CVAE's functional error is not uniform across the design space. As shown in the heatmap in Figure~\ref{fig:performance-heatmap}, errors are highest in low-friction designs, especially when the number of asperities is large. In this regime, it is difficult to resolve the contribution of each individual asperity, leading to high functional error. Conversely, in high-friction designs with many asperities, individual parameter errors tend to average out, yielding lower sMAPE. This averaging effect benefits high-asperity, high-friction designs, while surfaces with few asperities remain highly sensitive to small parameter deviations, resulting in higher relative errors.


\textbf{Sensitivity Analysis.}~Figure~\ref{fig:sensitivity_analysis} (Appendix~\ref{appendix:sensitivity}) that for surfaces with few asperities (e.g., 100), a 5\% perturbation in any parameter, including GMM weights \(w_1, w_2, w_3\) and standard deviations such as \(\sigma_{h1}, \sigma_{r4}\), induces substantial functional sMAPE variations, as all parameters contribute significantly. In contrast, for surfaces with many asperities (e.g., 10,000), variations in height and radius standard deviations are largely averaged out, leaving GMM weights as the dominant source of error and resulting in lower overall sMAPE. High-friction, high-asperity designs benefit from this \emph{averaging effect}, where errors in individual asperities partially cancel, whereas low-asperity surfaces remain highly sensitive, yielding higher relative errors.

\begin{figure}[b!]
    \centering
    \includegraphics[width = 0.75\textwidth]{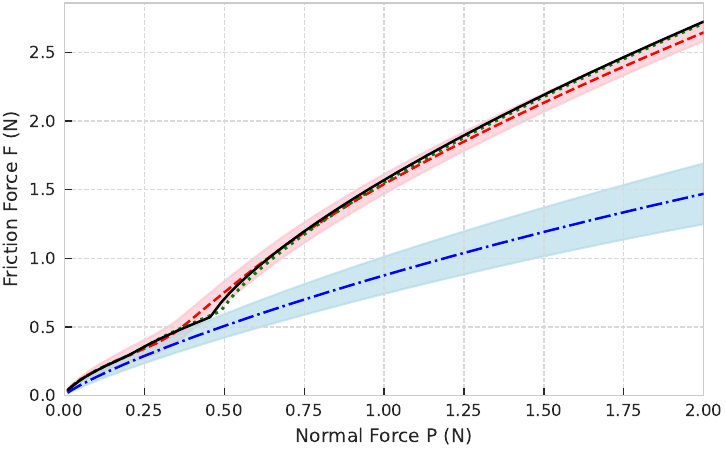}
    \caption{
        Comparison of friction laws generated by the VAE\,+\,CMA-ES and the CVAE models against the experimental friction law target (solid black line) and its GMM approximation (dotted green line). The result from the best-performing of 100 VAE\,+\,CMA-ES optimization runs is shown as a red dashed line. The surrounding pink shaded area represents the model uncertainty, quantified as one standard deviation across all 100 runs. For the CVAE, the blue dash-dotted line is the mean prediction over 100,000 inferences, with the light blue area indicating the corresponding generative standard deviation (mean ± 1 std. dev.).
    }
    \label{fig:experimental_VAE_CVAE_comp}
\end{figure}

\textbf{Performance on Out-of-Distribution Cases.} This performance gap is particularly pronounced in challenging, out-of-distribution (OOD) cases. Some experimental surfaces yield non-differentiable piecewise-nonlinear friction laws, which are rare (1–2\% of the dataset) and act as near-OOD samples relative to the smooth laws that dominate training. Example laws illustrating the different types present in the dataset are shown in Figure~\ref{fig:specific-samples-appendix} (Appendix~\ref{appendix:dataset}). The test case shown in Figure~\ref{fig:experimental_VAE_CVAE_comp} represents an even greater challenge: a truly out-of-distribution law derived from experimental observations~\cite{Aymard2023Thesis}, which was not seen during training and has some GMM parameters outside the bounds of the training dataset (see Appendix~\ref{appendix:dataset} for the boundaries). On this OOD target, the CVAE outputs only the dominant mode of its training data: a generic smooth friction law, completely ignoring the target's piecewise characteristic. In contrast, the VAE\,+\,CMA-ES approach, through its optimization-based search of the latent space, converges to a solution that accurately matches the target law. This comparison highlights a direct trade-off between inference speed and generalization. The CVAE provides rapid inference (0.15~s for 100,000 samples on a single RTX 4060 Ti 16 GB, see Figure~\ref{fig:cvae-convergence}) at the cost of robustness to OOD inputs. In contrast, VAE proves capable of generalizing to these challenging targets at the cost of several hours of optimization (\(\approx 100\)~s for a single run over 500 iterations initialized at the origin of the latent space, see Figures~\ref{fig:vae-convergence} and~\ref{fig:appendix-smape-grid}).

\section{Discussion}

\textbf{Absolute vs. Relative Performance in ML for Science}~Our results reveal a fundamental trade-off between inference speed and functional accuracy, with important implications for evaluating machine learning in scientific contexts. The CVAE performs amortized inference, generating candidate designs in milliseconds, a prerequisite for near-real-time control. However, this speed comes at the cost of reduced accuracy. Although the GMM parameters are estimated with near-perfect accuracy according to ML metrics (\(\approx 2.9\%\) sMAPE), the resulting error on the final friction law remains substantial (\(\approx 38\%\) sMAPE). The VAE, on the other hand, is able to reach a near-perfect accuracy on the final friction law (\(\approx 4.4\%\) sMAPE) and OOD generalization capabilities (\(\approx 2.8\%\) sMAPE) due to the inclusion of a latent space optimizer, but takes minutes to hours to converge to the final friction law.
This discrepancy serves as a stark case study. The common practice in ML is to assess the \textit{relative performance} of models, where a new architecture is deemed successful if it improves upon a benchmark, even by a small margin. However, for scientific applications, \textit{absolute performance} is paramount. We argue that to bridge the gap between ML research and practical science, we must shift our focus from relative rankings to absolute, physically meaningful error metrics aligned with the requirements of the target application.

\textbf{Practical Utility of Amortized Inference.}~Despite its limitations in functional precision, the CVAE remains a powerful tool for specific tasks. For exploratory design, its ability to generate a diverse ensemble of candidate solutions is valuable as it helps to answer whether a target friction law is physically achievable at all. Its ability to quantify uncertainty further guides decision-making, with narrow uncertainty bands indicating robust solutions, while wide bands may reveal targets near physical or manufacturability limits. For near-real-time control, the CVAE's value lies in providing rapid, "good enough" suggestions. Rather than perfectly matching a target, it can propose a standard, physically valid law that best approximates the desired behavior, which can then be adapted on-the-fly. This positions the CVAE as a practical near-real-time heuristic, balancing functional fidelity with computational constraints. Future work could investigate the development of hybrid inference that combines the strengths of both approaches: using the CVAE to generate a high-quality initial guess, followed by a few steps of a latent optimizer to refine the solution, potentially achieving both high speed and high accuracy.

\textbf{Broader Implications.}~The discrepancy we identify between parameter accuracy and functional fidelity is not unique to tribology; it illustrates a fundamental challenge in scientific inverse design. This challenge arises whenever a model is trained to predict an intermediate representation of a system, from which the final performance is then computed, a strategy found across numerous domains. In topology optimization, for example, the design is often represented by a density field that is filtered to yield the final topology~\cite{bourdin2001filters}. In molecular design, molecules are represented as graphs or strings, which must then be decoded into a 3D structure and evaluated with computationally expensive molecular simulations~\cite{sanchez2018inverse}. This reliance on intermediate representations is sometimes nested. In photonics, for example, the design of a large metasurface is often reduced to optimizing the geometry of a single, repeating unit cell~\cite{ji2023recent}. The inverse design model may not even predict these geometric parameters directly, but rather a low-dimensional latent vector used to generate them, introducing another intermediate step in the design process.
In each of these cases, small errors in the intermediate representation can cause large deviations in the final outcome. Our findings therefore argue that for ML to be a reliable tool for science, we must change how we measure success. Evaluation must shift from only accurately predicting an intermediate representation to making sure that it achieves absolute accuracy on the final objective.

\clearpage

\newpage

{
\small
\bibliographystyle{unsrtnat}
\bibliography{biblio_cvae_final.bib}

\begin{thebibliography}{43}
\providecommand{\natexlab}[1]{#1}
\providecommand{\url}[1]{\texttt{#1}}
\expandafter\ifx\csname urlstyle\endcsname\relax
  \providecommand{\doi}[1]{doi: #1}\else
  \providecommand{\doi}{doi: \begingroup \urlstyle{rm}\Url}\fi

\bibitem[Aymard et~al.(2024)Aymard, Delplanque, Dalmas, and Scheibert]{Aymard2024}
Antoine Aymard, Emilie Delplanque, Davy Dalmas, and Julien Scheibert.
\newblock Designing metainterfaces with specified friction laws.
\newblock \emph{Science}, 383\penalty0 (6679):\penalty0 200--204, 2024.
\newblock \doi{10.1126/science.adk4234}.

\bibitem[Aymard(2023)]{Aymard2023Thesis}
Antoine Aymard.
\newblock \emph{{Design et r{\'e}alisation d'interfaces textur{\'e}es {\'e}lastom{\'e}riques {\`a} loi de frottement pilot{\'e}e}}.
\newblock Thesis, {Ecole centrale de Lyon}, 2023.
\newblock URL \url{https://theses.hal.science/tel-04072503}.

\bibitem[Murarash et~al.(2011)Murarash, Itovich, and Varenberg]{murarash2011tuning}
Boris Murarash, Yan Itovich, and Michael Varenberg.
\newblock Tuning elastomer friction by hexagonal surface patterning.
\newblock \emph{Soft Matter}, 7\penalty0 (12):\penalty0 5553--5557, 2011.
\newblock \doi{10.1039/C1SM00015B}.

\bibitem[Li et~al.(2016)Li, Xu, Liu, Wang, and Liu]{li2016tuning}
Ning Li, Erjiang Xu, Ze~Liu, Xinyun Wang, and Lin Liu.
\newblock Tuning apparent friction coefficient by controlled patterning bulk metallic glasses surfaces.
\newblock \emph{Scientific Reports}, 6:\penalty0 39388, 2016.
\newblock \doi{10.1038/srep39388}.

\bibitem[Yu et~al.(2023)Yu, Hannah, Peng, Lin, Bhouri, Gupta, L\"{u}tjens, Will, Behrens, Busecke, Loose, Stern, Beucler, Harrop, Hillman, Jenney, Ferretti, Liu, Anandkumar, Brenowitz, Eyring, Geneva, Gentine, Mandt, Pathak, Subramaniam, Vondrick, Yu, Zanna, Zheng, Abernathey, Ahmed, Bader, Baldi, Barnes, Bretherton, Caldwell, Chuang, Han, HUANG, Iglesias-Suarez, Jantre, Kashinath, Khairoutdinov, Kurth, Lutsko, Ma, Mooers, Neelin, Randall, Shamekh, Taylor, Urban, Yuval, Zhang, and Pritchard]{yu2023climsim}
Sungduk Yu, Walter Hannah, Liran Peng, Jerry Lin, Mohamed~Aziz Bhouri, Ritwik Gupta, Bj\"{o}rn L\"{u}tjens, Justus~C. Will, Gunnar Behrens, Julius Busecke, Nora Loose, Charles Stern, Tom Beucler, Bryce Harrop, Benjamin Hillman, Andrea Jenney, Savannah~L. Ferretti, Nana Liu, Animashree Anandkumar, Noah Brenowitz, Veronika Eyring, Nicholas Geneva, Pierre Gentine, Stephan Mandt, Jaideep Pathak, Akshay Subramaniam, Carl Vondrick, Rose Yu, Laure Zanna, Tian Zheng, Ryan Abernathey, Fiaz Ahmed, David Bader, Pierre Baldi, Elizabeth Barnes, Christopher Bretherton, Peter Caldwell, Wayne Chuang, Yilun Han, YU~HUANG, Fernando Iglesias-Suarez, Sanket Jantre, Karthik Kashinath, Marat Khairoutdinov, Thorsten Kurth, Nicholas Lutsko, Po-Lun Ma, Griffin Mooers, J.~David Neelin, David Randall, Sara Shamekh, Mark Taylor, Nathan Urban, Janni Yuval, Guang Zhang, and Mike Pritchard.
\newblock Clim{S}im: A large multi-scale dataset for hybrid physics-{M}{L} climate emulation.
\newblock In A.~Oh, T.~Naumann, A.~Globerson, K.~Saenko, M.~Hardt, and S.~Levine, editors, \emph{Advances in Neural Information Processing Systems}, volume~36, pages 22070--22084. Curran Associates, Inc., 2023.
\newblock URL \url{https://proceedings.neurips.cc/paper_files/paper/2023/file/45fbcc01349292f5e059a0b8b02c8c3f-Paper-Datasets_and_Benchmarks.pdf}.

\bibitem[Zhu et~al.(2022)Zhu, Zhang, Zhang, and Li]{zhu2022generative}
Bao Zhu, Wenxin Zhang, Weisheng Zhang, and Hongxia Li.
\newblock Generative design of texture for sliding surface based on machine learning.
\newblock \emph{Tribology International}, 178:\penalty0 108139, 2022.
\newblock \doi{10.1016/j.triboint.2022.108139}.
\newblock URL \url{https://doi.org/10.1016/j.triboint.2022.108139}.

\bibitem[Kaliafetis et~al.(2025)Kaliafetis, Ardah, Ewen, and Dini]{kaliafetis2025using}
Filimonas Kaliafetis, Suhaib Ardah, James~P. Ewen, and Daniele Dini.
\newblock Using artificial neural networks to accelerate thermo-elastohydrodynamic lubrication simulations.
\newblock \emph{Tribology International}, 200:\penalty0 110978, 2025.
\newblock \doi{10.1016/j.triboint.2025.110978}.
\newblock URL \url{https://doi.org/10.1016/j.triboint.2025.110978}.

\bibitem[Karras et~al.(2021)Karras, Aittala, Laine, H\"{a}rk\"{o}nen, Hellsten, Lehtinen, and Aila]{karras_stylegan3}
Tero Karras, Miika Aittala, Samuli Laine, Erik H\"{a}rk\"{o}nen, Janne Hellsten, Jaakko Lehtinen, and Timo Aila.
\newblock Alias-free generative adversarial networks.
\newblock In M.~Ranzato, A.~Beygelzimer, Y.~Dauphin, P.S. Liang, and J.~Wortman Vaughan, editors, \emph{Advances in Neural Information Processing Systems}, volume~34, pages 852--863. Curran Associates, Inc., 2021.
\newblock URL \url{https://proceedings.neurips.cc/paper_files/paper/2021/file/076ccd93ad68be51f23707988e934906-Paper.pdf}.

\bibitem[Sauer et~al.(2023)Sauer, Karras, Laine, Geiger, and Aila]{sauer_stylegant_2023}
Axel Sauer, Tero Karras, Samuli Laine, Andreas Geiger, and Timo Aila.
\newblock {S}tyle{G}{A}{N}-{T}: {U}nlocking the {P}ower of {G}{A}{N}s for {F}ast {L}arge-{S}cale {T}ext-to-{I}mage {S}ynthesis.
\newblock In Andreas Krause, Emma Brunskill, Kyunghyun Cho, Barbara Engelhardt, Sivan Sabato, and Jonathan Scarlett, editors, \emph{International Conference on Machine Learning, {ICML} 2023, 23-29 July 2023, Honolulu, Hawaii, {USA}}, volume 202 of \emph{Proceedings of Machine Learning Research}, pages 30105--30118. {PMLR}, 2023.
\newblock URL \url{https://proceedings.mlr.press/v202/sauer23a.html}.

\bibitem[Liu et~al.(2018)Liu, Zhu, Rodrigues, Lee, and Cai]{luiphotonicmetasurface}
Zhaocheng Liu, Dayu Zhu, Sean~P. Rodrigues, Kyu-Tae Lee, and Wenshan Cai.
\newblock Generative model for the inverse design of metasurfaces.
\newblock \emph{Nano Letters}, 18\penalty0 (10):\penalty0 6570--6576, 2018.
\newblock \doi{10.1021/acs.nanolett.8b03171}.

\bibitem[Hen et~al.(2025)Hen, Yosef, Raviv, Giryes, and Scheuer]{acsphotonics2025diffusion}
Liav Hen, Erez Yosef, Dan Raviv, Raja Giryes, and Jacob Scheuer.
\newblock Inverse design of diffractive metasurfaces using diffusion models.
\newblock \emph{ACS Photonics}, 2025.
\newblock \doi{10.1021/acsphotonics.5c01384}.
\newblock URL \url{https://doi.org/10.1021/acsphotonics.5c01384}.

\bibitem[Sanchez-Lengeling and Aspuru-Guzik(2018)]{sanchez2018inverse}
Benjamin Sanchez-Lengeling and Alán Aspuru-Guzik.
\newblock Inverse molecular design using machine learning: Generative models for matter engineering.
\newblock \emph{Science}, 361\penalty0 (6400):\penalty0 360--365, 2018.
\newblock \doi{10.1126/science.aat2663}.

\bibitem[Bastek and Kochmann(2023)]{bastek2023inverse}
Jan-Hendrik Bastek and Dennis~M. Kochmann.
\newblock Inverse design of nonlinear mechanical metamaterials via video denoising diffusion models.
\newblock \emph{Nature Machine Intelligence}, 5:\penalty0 1466--1475, 2023.
\newblock \doi{10.1038/s42256-023-00762-x}.
\newblock URL \url{https://doi.org/10.1038/s42256-023-00762-x}.
\newblock Published: 11 December 2023.

\bibitem[Brzin and Brojan(2024)]{brzin2024}
Toma{\v{z}} Brzin and Miha Brojan.
\newblock Using a generative adversarial network for the inverse design of soft morphing composite beams.
\newblock \emph{Engineering Applications of Artificial Intelligence}, 133:\penalty0 108527, July 2024.
\newblock \doi{10.1016/j.engappai.2024.108527}.
\newblock URL \url{https://doi.org/10.1016/j.engappai.2024.108527}.

\bibitem[Higgins et~al.(2017)Higgins, Matthey, Pal, Burgess, Glorot, Botvinick, Mohamed, and Lerchner]{higgins2017beta-kl}
Irina Higgins, Lo{\"{\i}}c Matthey, Arka Pal, Christopher~P. Burgess, Xavier Glorot, Matthew~M. Botvinick, Shakir Mohamed, and Alexander Lerchner.
\newblock beta-{V}{A}{E}: {Learning} {Basic} {Visual} {Concepts} with a {Constrained} {Variational} {Framework}.
\newblock In \emph{5th International Conference on Learning Representations, {ICLR} 2017, Toulon, France, April 24-26, 2017, Conference Track Proceedings}, 2017.
\newblock URL \url{https://openreview.net/forum?id=Sy2fzU9gl}.

\bibitem[Tomczak and Welling(2018)]{tomczak2018vaevampprior}
Jakub~M. Tomczak and Max Welling.
\newblock Vae with a vampprior, 2018.
\newblock URL \url{https://arxiv.org/abs/1705.07120}.

\bibitem[Zhao et~al.(2024)Zhao, Zhang, Sun, Yang, Chen, and Wang]{zhao2024ccdpm}
Yanxuan Zhao, Peng Zhang, Guopeng Sun, Zhigong Yang, Jianqiang Chen, and Yueqing Wang.
\newblock Cc{D}{P}{M}: {A} {Continuous} {Conditional} {Diffusion} {Probabilistic} {Model} for {Inverse} {Design}.
\newblock \emph{Proceedings of the AAAI Conference on Artificial Intelligence}, 38\penalty0 (15):\penalty0 17033--17041, 2024.
\newblock \doi{10.1609/aaai.v38i15.29647}.

\bibitem[Pan et~al.(2022)Pan, Niu, and Zhang]{pan2022unigan}
Ziqi Pan, Li~Niu, and Liqing Zhang.
\newblock Uni{G}{A}{N}: Reducing {Mode} {Collapse} in {G}{A}{N}s using a {Uniform} {Generator}.
\newblock In S.~Koyejo, S.~Mohamed, A.~Agarwal, D.~Belgrave, K.~Cho, and A.~Oh, editors, \emph{Advances in Neural Information Processing Systems}, volume~35, pages 37690--37703. Curran Associates, Inc., 2022.
\newblock URL \url{https://proceedings.neurips.cc/paper_files/paper/2022/file/f5537b8d8fd126c7fe9d7429b181b1eb-Paper-Conference.pdf}.

\bibitem[Renda et~al.(2020)Renda, Chen, Mendis, and Carbin]{Renda2020DiffTune}
Alex Renda, Yishen Chen, Charith Mendis, and Michael Carbin.
\newblock Diff{T}une: {Optimizing} {C}{P}{U} {Simulator} {Parameters} with {Learned} {Differentiable} {Surrogates}.
\newblock In \emph{2020 53rd Annual IEEE/ACM International Symposium on Microarchitecture (MICRO)}, pages 442--455, 2020.
\newblock \doi{10.1109/MICRO50266.2020.00045}.

\bibitem[Huang et~al.(2025)Huang, Liu, Wu, Lin, Tan, Zhang, Gao, Li, Liu, Liu, Wu, and Li]{huang2025eva}
Yufei Huang, Yunshu Liu, Lirong Wu, Haitao Lin, Cheng Tan, Odin Zhang, Zhangyang Gao, Siyuan Li, Zicheng Liu, Yunfan Liu, Tailin Wu, and Stan~Z Li.
\newblock {EVA}: {Geometric} {Inverse} {Design} for {Fast} {Motif}-{Scaffolding} with {Coupled} {Flow}.
\newblock In Y.~Yue, A.~Garg, N.~Peng, F.~Sha, and R.~Yu, editors, \emph{International Conference on Representation Learning}, volume 2025, pages 31714--31733, 2025.
\newblock URL \url{https://proceedings.iclr.cc/paper_files/paper/2025/file/4eb2c0adafbe71269f3a772c130f9e53-Paper-Conference.pdf}.

\bibitem[Huang et~al.(2022)Huang, Hong, Wawrzynek, Subedar, and Shao]{huang2022learning}
Qijing Huang, Charles Hong, John Wawrzynek, Mahesh Subedar, and Yakun~Sophia Shao.
\newblock Learning {A} {Continuous} and {Reconstructible} {Latent} {Space} for {Hardware} {Accelerator} {Design}.
\newblock In \emph{2022 IEEE International Symposium on Performance Analysis of Systems and Software (ISPASS)}, pages 277--287, 2022.
\newblock \doi{10.1109/ISPASS55109.2022.00041}.

\bibitem[Ren et~al.(2025)Ren, Esfahani, and Georghiou]{ren2025inverse}
Ke~Ren, Peyman~Mohajerin Esfahani, and Angelos Georghiou.
\newblock {Inverse} {Optimization} via {Learning} {Feasible} {Regions}.
\newblock In \emph{Proceedings of the 42nd {International} {Conference} on {Machine} {Learning}}. PMLR, 2025.
\newblock URL \url{https://openreview.net/pdf?id=lEV0x6aDKc}.
\newblock To appear.

\bibitem[Coley et~al.(2018)Coley, Rogers, Green, and Jensen]{Coley2018SCScore}
Connor~W. Coley, Luke Rogers, William~H. Green, and Klavs~F. Jensen.
\newblock {SCScore}: {Synthetic} {Complexity} {Learned} from a {Reaction} {Corpus}.
\newblock \emph{Journal of Chemical Information and Modeling}, 58\penalty0 (2):\penalty0 252--261, 2018.
\newblock \doi{10.1021/acs.jcim.7b00622}.

\bibitem[Thakkar et~al.(2021)Thakkar, Chadimová, Bjerrum, Engkvist, and Reymond]{Thakkar2021RAscore}
Amol Thakkar, Veronika Chadimová, Esben~Jannik Bjerrum, Ola Engkvist, and Jean-Louis Reymond.
\newblock Retrosynthetic accessibility score ({R}{A}score) – rapid machine learned synthesizability classification from {A}{I}-driven retrosynthetic planning.
\newblock \emph{Chemical Science}, 12:\penalty0 3339--3349, 2021.
\newblock \doi{10.1039/D0SC05401A}.

\bibitem[Daras et~al.(2024)Daras, Nie, Kreis, Dimakis, Mardani, Kovachki, and Vahdat]{daras2024warped}
Giannis Daras, Weili Nie, Karsten Kreis, Alexandros~G. Dimakis, Morteza Mardani, Nikola~B. Kovachki, and Arash Vahdat.
\newblock {Warped} {Diffusion}: {Solving} {Video} {Inverse} {Problems} with {Image} {Diffusion} {Models}.
\newblock In A.~Globerson, L.~Mackey, D.~Belgrave, A.~Fan, U.~Paquet, J.~Tomczak, and C.~Zhang, editors, \emph{Advances in Neural Information Processing Systems}, volume~37, pages 101116--101143. Curran Associates, Inc., 2024.
\newblock URL \url{https://proceedings.neurips.cc/paper_files/paper/2024/file/b736c4b0b38876c9249db9bd900c1a86-Paper-Conference.pdf}.

\bibitem[Khaireh-Walieh et~al.(2023)Khaireh-Walieh, Langevin, Bennet, Teytaud, Moreau, and Wiecha]{khaireh2023newcomer}
Abdourahman Khaireh-Walieh, Denis Langevin, Pauline Bennet, Olivier Teytaud, Antoine Moreau, and Peter~R. Wiecha.
\newblock A newcomer’s guide to deep learning for inverse design in nano-photonics.
\newblock \emph{Nanophotonics}, 12\penalty0 (24):\penalty0 4387--4414, 2023.
\newblock \doi{10.1515/nanoph-2023-0527}.

\bibitem[Sauer et~al.(2022)Sauer, Schwarz, and Geiger]{sauer_styleganxl}
Axel Sauer, Katja Schwarz, and Andreas Geiger.
\newblock Stylegan-xl: Scaling stylegan to large diverse datasets.
\newblock In \emph{ACM SIGGRAPH 2022 Conference Proceedings}, SIGGRAPH '22, New York, NY, USA, 2022. Association for Computing Machinery.
\newblock ISBN 9781450393379.
\newblock \doi{10.1145/3528233.3530738}.
\newblock URL \url{https://doi.org/10.1145/3528233.3530738}.

\bibitem[Greenwood and Williamson(1966)]{Greenwood1966}
J.~A. Greenwood and J.~B.~P. Williamson.
\newblock Contact of nominally flat surfaces.
\newblock \emph{Proceedings of the Royal Society of London. Series A. Mathematical and Physical Sciences}, 295\penalty0 (1442):\penalty0 300--319, 1966.
\newblock \doi{10.1098/rspa.1966.0242}.

\bibitem[Silva et~al.(2024)Silva, Costa, Luz, Oliveira, and Profito]{Silva2024}
Samuel A~N Silva, Henara~L Costa, Felipe K~C Luz, Elton Y~G Oliveira, and Francisco~J Profito.
\newblock Applying {Gaussian} mixture models for enhanced characterization of featured surfaces and mixed lubrication analysis.
\newblock \emph{Surface Topography: Metrology and Properties}, 12\penalty0 (3):\penalty0 035016, jul 2024.
\newblock \doi{10.1088/2051-672X/ad4571}.

\bibitem[Sobol'(1967)]{sobol1967distribution}
Ilya~M. Sobol'.
\newblock On the distribution of points in a cube and the approximate evaluation of integrals.
\newblock \emph{USSR Computational Mathematics and Mathematical Physics}, 7\penalty0 (4):\penalty0 86--112, 1967.
\newblock \doi{10.1016/0041-5553(67)90144-9}.

\bibitem[Kingma and Welling(2014)]{kingma2013vae}
Diederik~P. Kingma and Max Welling.
\newblock Auto-{Encoding} {Variational} {Bayes}.
\newblock In Yoshua Bengio and Yann~Le Cun, editors, \emph{2nd International Conference on Learning Representations, {ICLR} 2014, Banff, AB, Canada, April 14-16, 2014, Conference Track Proceedings}, 2014.
\newblock URL \url{http://arxiv.org/abs/1312.6114}.

\bibitem[Huber(1964)]{huber1964robust}
Peter~J. Huber.
\newblock {Robust Estimation of a Location Parameter}.
\newblock \emph{The Annals of Mathematical Statistics}, 35\penalty0 (1):\penalty0 73--101, 1964.
\newblock \doi{10.1214/aoms/1177703732}.

\bibitem[Akiba et~al.(2019)Akiba, Sano, Yanase, Ohta, and Koyama]{Akiba2019Optuna}
Takuya Akiba, Shotaro Sano, Toshihiko Yanase, Takeru Ohta, and Masanori Koyama.
\newblock Optuna: A {Next}-generation {Hyperparameter} {Optimization} {Framework}.
\newblock In \emph{Proceedings of the 25th ACM SIGKDD International Conference on Knowledge Discovery \& Data Mining}, KDD '19, page 2623–2631, New York, NY, USA, 2019. Association for Computing Machinery.
\newblock \doi{10.1145/3292500.3330701}.

\bibitem[Chen and Guestrin(2016)]{chen2016xgboost}
Tianqi Chen and Carlos Guestrin.
\newblock {XGBoost: A Scalable Tree Boosting System}.
\newblock In \emph{Proceedings of the 22nd ACM SIGKDD International Conference on Knowledge Discovery and Data Mining}, KDD '16, page 785–794, New York, NY, USA, 2016. Association for Computing Machinery.
\newblock \doi{10.1145/2939672.2939785}.

\bibitem[Hansen(2016)]{hansen2006cma}
Nikolaus Hansen.
\newblock The {CMA} {Evolution} {Strategy}: {A} {Tutorial}.
\newblock 2016.
\newblock URL \url{http://arxiv.org/abs/1604.00772}.

\bibitem[Bourdin(2001)]{bourdin2001filters}
Blaise Bourdin.
\newblock Filters in topology optimization.
\newblock \emph{International Journal for Numerical Methods in Engineering}, 50\penalty0 (9):\penalty0 2143--2158, 2001.
\newblock \doi{10.1002/nme.116}.

\bibitem[Ji et~al.(2023)Ji, Chang, Xu, Gao, Gr{\"o}blacher, Urbach, and Adam]{ji2023recent}
Wenye Ji, Jin Chang, He-Xiu Xu, Jian~Rong Gao, Simon Gr{\"o}blacher, H.~Paul Urbach, and Aur{\`e}le J.~L. Adam.
\newblock Recent advances in metasurface design and quantum optics applications with machine learning, physics-informed neural networks, and topology optimization methods.
\newblock \emph{Light: Science \& Applications}, 12\penalty0 (1):\penalty0 169, 2023.
\newblock \doi{10.1038/s41377-023-01218-y}.

\bibitem[Bradbury et~al.(2018)Bradbury, Frostig, Hawkins, Johnson, Leary, Maclaurin, van~den Driessche, Wanderman-Milne, and Quinnradical]{jax2018github}
James Bradbury, Roy Frostig, Peter Hawkins, Matthew~James Johnson, Chris Leary, Dougal Maclaurin, George van~den Driessche, Skye Wanderman-Milne, and Quinnradical.
\newblock {JAX}: composable transformations of {P}ython+{N}um{P}y programs, 2018.
\newblock URL \url{http://github.com/google/jax}.

\bibitem[Paszke et~al.(2019)Paszke, Gross, Massa, Lerer, Bradbury, Chanan, Killeen, Lin, Gimelshein, Antiga, Desmaison, Kopf, Yang, DeVito, Raison, Tejani, Chilamkurthy, Steiner, Fang, Bai, and Chintala]{paszke2019pytorch}
Adam Paszke, Sam Gross, Francisco Massa, Adam Lerer, James Bradbury, Gregory Chanan, Trevor Killeen, Zeming Lin, Natalia Gimelshein, Luca Antiga, Alban Desmaison, Andreas Kopf, Edward Yang, Zachary DeVito, Martin Raison, Alykhan Tejani, Sasank Chilamkurthy, Benoit Steiner, Lu~Fang, Junjie Bai, and Soumith Chintala.
\newblock Py{T}orch: {An} {Imperative} {Style}, {High}-{Performance} {Deep} {Learning} {Library}.
\newblock In H.~Wallach, H.~Larochelle, A.~Beygelzimer, F.~d\textquotesingle Alch\'{e}-Buc, E.~Fox, and R.~Garnett, editors, \emph{Advances in Neural Information Processing Systems}, volume~32, pages 8024--8035. Curran Associates, Inc., 2019.
\newblock URL \url{https://proceedings.neurips.cc/paper_files/paper/2019/file/bdbca288fee7f92f2bfa9f7012727740-Paper.pdf}.

\bibitem[Pedregosa et~al.(2011)Pedregosa, Varoquaux, Gramfort, Michel, Thirion, Grisel, Blondel, Prettenhofer, Weiss, Dubourg, Vanderplas, Passos, Cournapeau, Brucher, Perrot, and {{\'E}}douard Duchesnay]{scikit-learn}
Fabian Pedregosa, Ga{{\"e}}l Varoquaux, Alexandre Gramfort, Vincent Michel, Bertrand Thirion, Olivier Grisel, Mathieu Blondel, Peter Prettenhofer, Ron Weiss, Vincent Dubourg, Jake Vanderplas, Alexandre Passos, David Cournapeau, Matthieu Brucher, Matthieu Perrot, and {{\'E}}douard Duchesnay.
\newblock Scikit-learn: Machine learning in python.
\newblock \emph{Journal of Machine Learning Research}, 12\penalty0 (85):\penalty0 2825--2830, 2011.
\newblock URL \url{http://jmlr.org/papers/v12/pedregosa11a.html}.

\bibitem[{W}es {M}c{K}inney(2010)]{mckinney-proc-scipy-2010}
{W}es {M}c{K}inney.
\newblock {D}ata {S}tructures for {S}tatistical {C}omputing in {P}ython.
\newblock In {S}t\'efan van~der {W}alt and {J}arrod {M}illman, editors, \emph{{P}roceedings of the 9th {P}ython in {S}cience {C}onference}, pages 56 -- 61, 2010.
\newblock \doi{10.25080/Majora-92bf1922-00a}.

\bibitem[Hunter(2007)]{Hunter2007}
John~D. Hunter.
\newblock {Matplotlib: A 2D Graphics Environment}.
\newblock \emph{Computing in Science \& Engineering}, 9\penalty0 (3):\penalty0 90--95, 2007.
\newblock \doi{10.1109/MCSE.2007.55}.

\bibitem[Waskom(2021)]{Waskom2021_seaborn}
Michael~L. Waskom.
\newblock seaborn: statistical data visualization.
\newblock \emph{Journal of Open Source Software}, 6\penalty0 (60):\penalty0 3021, 2021.
\newblock \doi{10.21105/joss.03021}.

\end{thebibliography}
}


\newpage
\appendix

\section{Dataset}
\label{appendix:dataset}

\subsection{Surface Topography Modeling}

Unlike the classic GW model, which assumes simple, independent distributions for asperity properties (i.e., an exponential height distribution and a constant radius of curvature), we model the joint distribution of asperity height \(h\) and radius of curvature \(R\) using a Gaussian Mixture Model (GMM). This approach allows for the representation of complex, multi-modal surface topographies~\cite{Silva2024}. The joint probability density function \(\Phi(h, R)\) for a population of \(N\) asperities is described by a GMM with \(K\) components:
\begin{equation}
\Phi(h, R | \bm{\theta}) = \sum_{k=1}^{K} w_k \, \mathcal{N}\left( \begin{pmatrix} h \\ R \end{pmatrix} \bigg| \bm{\mu}_k, \bm{\Sigma}_k \right)
\label{eq:gmm}
\end{equation}
where:
\begin{itemize}
    \item \(K\) is the number of Gaussian components (in this work, \(K=4\)).
    \item \(w_k\) are the mixture weights, satisfying \(w_k \ge 0\) and \(\sum_{k=1}^{K} w_k = 1\).
    \item \(\mathcal{N}(\cdot | \bm{\mu}_k, \bm{\Sigma}_k)\) is a bivariate Gaussian probability density function.
    \item \(\bm{\theta} = \{w_k, \bm{\mu}_k, \bm{\Sigma}_k\}_{k=1}^K\) represents the complete set of GMM parameters.
\end{itemize}

The mean vector \(\bm{\mu}_k\) and the covariance matrix \(\bm{\Sigma}_k\) for each component \(k\) are defined as:
\begin{equation}
\bm{\mu}_k = \begin{pmatrix} \mu_{h,k} \\ \mu_{R,k} \end{pmatrix}
\end{equation}

\begin{equation}
\bm{\Sigma}_k = \begin{pmatrix}
\sigma_{h,k}^2 & \rho_k \sigma_{h,k} \sigma_{R,k} \\
\rho_k \sigma_{h,k} \sigma_{R,k} & \sigma_{R,k}^2
\end{pmatrix}
\label{eq:cov_matrix}
\end{equation}
here, \(\mu_{h,k}\) and \(\mu_{R,k}\) are the mean height and radius, \(\sigma_{h,k}\) and \(\sigma_{R,k}\) are their standard deviations, and \(\rho_k\) is the correlation coefficient between height and radius for the \(k\)-th component. For \(K=4\), the complete topography model is defined by 23 parameters. Empirical results indicate that choosing four components provides sufficient flexibility to represent diverse friction laws, including the complex behaviors observed experimentally (Figure~\ref{fig:experimental_VAE_CVAE_comp}), without introducing unnecessary complexity.

\subsection{Theoretical Force Calculation}

Theoretically, for a surface defined by parameters \(\bm{\theta}\) with \(N\) asperities and a given indentation \(\delta\), the expected total normal force \(P(\delta)\) and friction force \(F(\delta)\) can be expressed as:
\begin{equation}
\left\{
\begin{array}{l}
\displaystyle
P(\delta) = N \iint_{\Omega} \frac{4}{3} E^* \sqrt{R}\,(h - \delta)^{3/2}\,\Phi(h, R | \bm{\theta})\, dR\, dh \\[12pt]
\displaystyle
F(\delta) = N \iint_{\Omega} \sigma B \pi R \,(h - \delta)\,\Phi(h, R | \bm{\theta})\, dR\, dh
\end{array}
\right.
\label{eq:integral_form}
\end{equation}
where \(E^*\) is the composite Young's modulus, \(\sigma\) is the interfacial shear strength, and the integration domain \(\Omega\) covers contacting asperities (\(h > \delta\)).

\subsection{Dataset Generation}

To generate a comprehensive dataset suitable for machine learning, we systematically explore the space of possible topographies and their scaling with asperity count. The generation process involves a nested loop structure.

\textbf{Parameter Space Exploration.}~To ensure a diverse and uniform coverage of surface types, we first sample the 23-dimensional GMM parameter space. A Sobol sequence is used to generate a low-discrepancy set of parameter vectors \(\{\bm{\theta}\}\). Each vector \(\bm{\theta}\) represents a unique "recipe" for a surface topography distribution.

\textbf{Asperity Count Sweep.}~For each GMM recipe \(\bm{\theta}\) generated in the outer loop, we perform a series of simulations, systematically varying the number of asperities, \(N\). We use a predefined, logarithmically-spaced set of values for \(N\), ranging from 30 to 12,000.

\textbf{Forward Simulation.}~For each pair of \((\bm{\theta}, N)\), we execute a direct simulation. This involves generating a discrete population of \(N\) asperities by drawing \((h_i, R_i)\) samples from the GMM defined by \(\bm{\theta}\). The total forces for a given indentation \(\delta\) are then calculated via a discrete summation, which serves as a Monte Carlo estimator of the integrals in Eq. \ref{eq:integral_form}:
\begin{equation}
P(\delta) = \sum_{i=1}^{N} \frac{4}{3} E^* \sqrt{R_i} \left( \max(0, h_i - \delta) \right)^{3/2}
\label{eq:summation_force}
\end{equation}

\begin{equation}
F(\delta) = \sum_{i=1}^{N} \sigma B \pi R_i \left( \max(0, h_i - \delta) \right)
\label{eq:summation_friction}
\end{equation}
By computing the pair \((P(\delta), F(\delta))\) over a range of indentations, we generate a single \(\bm{F}(P)\) friction law. This process yields a unique sample for each \((\bm{\theta}, N)\) pair (discretized in 128 values). In our final dataset, the GMM parameters \(\bm{\theta}\) serve as the target outputs, while the model inputs consist of the discretized friction curve and the corresponding number of asperities \(N\).

\begin{figure}[h!] 
    \centering
    
    \includegraphics{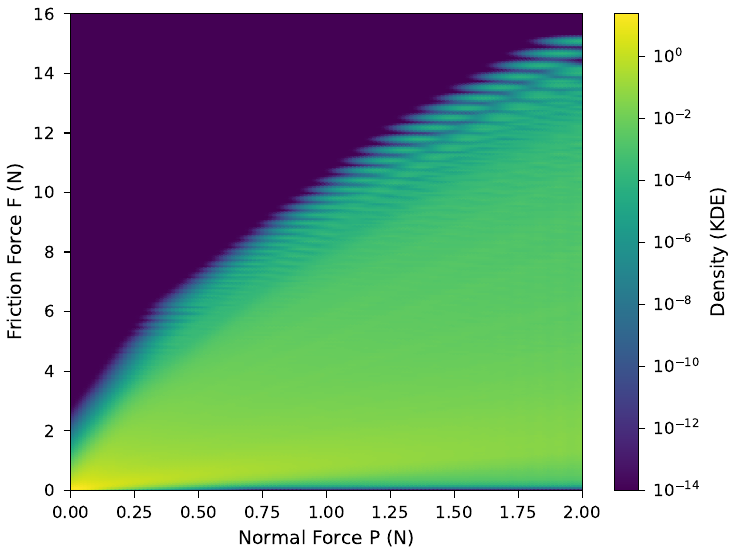}
    
    \caption{
        Two-dimensional probability density of the dataset, estimated via a GPU-accelerated Kernel Density Estimation (KDE). The optimal KDE hyperparameters (kernel type and bandwidth) were determined via 3-fold cross-validation on a random subset of 5,000 friction law curves. The final density was then computed on a larger sample of 16,384 curves using these optimal parameters. The x-axis represents the normal force \(P\), and the y-axis represents the corresponding friction force \(F\). The color intensity, plotted on a logarithmic scale, indicates the probability density, defining the valid physical domain for our generative models.
    }
    \label{fig:data-density-kde} 
\end{figure}

\textbf{Implementation Details.}~The entire simulation pipeline is implemented in Python, using the JAX library for its just-in-time (JIT) compilation and automatic vectorization capabilities. Computations were executed on a single Nvidia GeForce RTX 4060 Ti 16\,GB GPU. The use of JAX's \texttt{vmap} function enables the parallel processing of large simulation batches (in this work, 32,768 simulations per call). This high-throughput computational strategy is essential for the practical generation of the large-scale dataset within a feasible timeframe.

\textbf{Dataset Scale and Preprocessing.}~The entire data generation pipeline produced a final dataset of 200,278,016 samples in a total of 10 hours and 30 minutes. To prepare the data for model training and evaluation, the dataset was first randomly split into training (70\%), validation (15\%), and test~(15\%) sets. To prevent data leakage, a min-max scaler was then fitted exclusively on the training data. This scaler was subsequently used to normalize all input features (the discretized friction law and asperity count) and target GMM parameters across all three sets into a consistent range of \([-1, 1]\).

\subsection{Dataset Boundaries}

This section details the specific boundaries and parameters used to generate the 200-million-sample dataset. The generation process involved two main loops: an outer loop that sampled the 23-dimensional Gaussian Mixture Model (GMM) parameter space using a Sobol sequence, and an inner loop that, for each sampled GMM recipe, systematically swept through a predefined set of asperity counts (\(N\)).

The core physical constants, based on a Polydimethylsiloxane (PDMS)-on-glass interface, and other fixed simulation parameters are listed in Table~\ref{tab:physics_params}.

\begin{table}[h!]
\centering
\caption{Fixed physical and geometric parameters used in the forward simulation.}
\label{tab:physics_params}
\begin{tabular}{llc}
\toprule
\textbf{Parameter} & \textbf{Description} & \textbf{Value} \\
\midrule
$E^*$ & Composite elastic modulus & 1.36 MPa \\
$\sigma$ & Interfacial shear strength & 0.40 MPa \\
$B$ & Area reduction ratio (taken from~\cite{Aymard2024}) & 0.85 \\
$h_{bounds}$ & Absolute height bounds for asperities & [0.0, 300.0] µm \\
$R_{bounds}$ & Absolute radius bounds for asperities & [10.0, 600.0] µm \\
\bottomrule
\end{tabular}
\end{table}

The 23-dimensional hypercube for the GMM parameter sampling was defined by the bounds listed in Table~\ref{tab:gmm_bounds}. For means, standard deviations, and correlations, the bounds were identical across all four Gaussian components (\(k \in \{1, 2, 3, 4\}\)), allowing for a compact representation. The mixture weights (\(w_1, w_2, w_3\)) were sampled such that their sum was less than or equal to 1, with \(w_4\) being derived as~\(1 - \sum_{i=1}^{3} w_i\).

\begin{table}[h!]
\centering
\caption{Parameter space boundaries for the 4-component GMM Sobol sampling.}
\label{tab:gmm_bounds}
\begin{tabular}{llc}
\toprule
\textbf{Parameter Type} & \textbf{Description} & \textbf{Bounds} \\
\midrule
$w_k$ & Mixture weights (for $k \in \{1,2,3\}$) & [0.0, 1.0] \\
$\mu_{h,k}$ & Mean of asperity height $h$ & [50.0, 250.0] µm \\
$\mu_{R,k}$ & Mean of asperity radius $R$ & [50.0, 500.0] µm \\
$\sigma_{h,k}$ & Std. dev. of asperity height $h$ & [10.0, 80.0] µm \\
$\sigma_{R,k}$ & Std. dev. of asperity radius $R$ & [10.0, 100.0] µm \\
$\rho_k$ & Correlation coefficient between $h$ and $R$ & [-0.9, 0.9] \\
\bottomrule
\end{tabular}
\end{table}

Finally, for each of the 12.5 million unique GMM recipes, a full friction law was computed for each asperity count. The raw simulation output was then processed to generate the final feature vector, as detailed in Table~\ref{tab:discretization_params}.

\begin{table}[h!]
\centering
\caption{Discretization parameters for asperity count and feature extraction.}
\label{tab:discretization_params}
\begin{tabular}{llc}
\toprule
\textbf{Parameter} & \textbf{Description} & \textbf{Value / Range} \\
\midrule
$N$ & Asperity counts simulated per GMM recipe & [30, 50, ..., 12000] (16 steps) \\
$P_{grid}$ & Standardized grid for normal force features & 128 points from 0.01 to 2.0 N \\
$\delta_{grid}$ & Indentation values for forward model & 256 points from 0.001 to 300.0 µm \\
\bottomrule
\end{tabular}
\end{table}

\subsection{Statistical Description}

This section provides a statistical summary of the unscaled dataset, calculated across all 200 million samples. The variables are divided into two groups: the input features (Table~\ref{tab:desc_stats_features}), which consist of the 128-point discretized friction law and the asperity count, and the target GMM parameters (Table~\ref{tab:desc_stats_targets}).

The statistics confirm the quasi-uniform nature of the sampled GMM parameter space, a direct result of using a Sobol sequence for generation. This is evidenced by the near-zero skewness and negative kurtosis (platykurtic distribution) for these parameters. In contrast, the friction law features \(F_i\) exhibit a progressively increasing positive skew, reflecting the physical constraints of the contact mechanics model where low-friction outcomes are more prevalent than high-friction ones.

{
\begin{longtable}{@{}lrrrrrr@{}}
  \caption{\normalsize Descriptive statistics for the input features (discretized friction law and asperity count).}
  \label{tab:desc_stats_features} \\
  [-1.5ex]
  \toprule
  Input Variable & Mean & Std Dev & Min & Max & Skewness & Kurtosis \\
  \midrule
  \endfirsthead

  \multicolumn{7}{c}%
  {{\bfseries \tablename\ \thetable{} -- continued from previous page}} \\
  \toprule
  Input Variable & Mean & Std Dev & Min & Max & Skewness & Kurtosis \\
  \midrule
  \endhead

  \midrule
  \multicolumn{7}{r@{}}{{Continued on next page}} \\
  \endfoot

  \bottomrule
  \endlastfoot

  \texttt{F\_0} & 0.0517 & 0.0316 & 0.0015 & 0.1346 & 0.5235 & -0.9088 \\
  \texttt{F\_1} & 0.0536 & 0.0329 & 0.0016 & 0.1402 & 0.5378 & -0.8924 \\
  \texttt{F\_2} & 0.0555 & 0.0343 & 0.0016 & 0.1460 & 0.5523 & -0.8757 \\
  \texttt{F\_3} & 0.0575 & 0.0357 & 0.0017 & 0.1522 & 0.5668 & -0.8584 \\
  \texttt{F\_4} & 0.0595 & 0.0372 & 0.0017 & 0.1587 & 0.5815 & -0.8403 \\
  \texttt{F\_5} & 0.0616 & 0.0388 & 0.0018 & 0.1654 & 0.5962 & -0.8214 \\
  \texttt{F\_6} & 0.0638 & 0.0404 & 0.0018 & 0.1725 & 0.6111 & -0.8017 \\
  \texttt{F\_7} & 0.0661 & 0.0420 & 0.0019 & 0.1798 & 0.6260 & -0.7814 \\
  \texttt{F\_8} & 0.0684 & 0.0438 & 0.0019 & 0.1875 & 0.6410 & -0.7603 \\
  \texttt{F\_9} & 0.0708 & 0.0456 & 0.0020 & 0.1955 & 0.6561 & -0.7383 \\
  \texttt{F\_10} & 0.0733 & 0.0474 & 0.0020 & 0.2038 & 0.6713 & -0.7155 \\
  \texttt{F\_11} & 0.0759 & 0.0494 & 0.0021 & 0.2125 & 0.6866 & -0.6924 \\
  \texttt{F\_12} & 0.0786 & 0.0514 & 0.0022 & 0.2211 & 0.7020 & -0.6679 \\
  \texttt{F\_13} & 0.0813 & 0.0535 & 0.0022 & 0.2305 & 0.7175 & -0.6426 \\
  \texttt{F\_14} & 0.0842 & 0.0556 & 0.0023 & 0.2400 & 0.7331 & -0.6164 \\
  \texttt{F\_15} & 0.0871 & 0.0579 & 0.0023 & 0.2502 & 0.7488 & -0.5897 \\
  \texttt{F\_16} & 0.0902 & 0.0602 & 0.0024 & 0.2609 & 0.7646 & -0.5621 \\
  \texttt{F\_17} & 0.0933 & 0.0626 & 0.0030 & 0.2720 & 0.7805 & -0.5336 \\
  \texttt{F\_18} & 0.0966 & 0.0652 & 0.0031 & 0.2836 & 0.7964 & -0.5038 \\
  \texttt{F\_19} & 0.0999 & 0.0678 & 0.0032 & 0.2957 & 0.8125 & -0.4731 \\
  \texttt{F\_20} & 0.1034 & 0.0705 & 0.0033 & 0.3083 & 0.8286 & -0.4419 \\
  \texttt{F\_21} & 0.1070 & 0.0733 & 0.0035 & 0.3214 & 0.8449 & -0.4097 \\
  \texttt{F\_22} & 0.1107 & 0.0762 & 0.0036 & 0.3345 & 0.8613 & -0.3764 \\
  \texttt{F\_23} & 0.1145 & 0.0792 & 0.0037 & 0.3487 & 0.8778 & -0.3420 \\
  \texttt{F\_24} & 0.1184 & 0.0824 & 0.0038 & 0.3636 & 0.8943 & -0.3067 \\
  \texttt{F\_25} & 0.1225 & 0.0856 & 0.0039 & 0.3791 & 0.9109 & -0.2703 \\
  \texttt{F\_26} & 0.1267 & 0.0890 & 0.0041 & 0.3938 & 0.9277 & -0.2329 \\
  \texttt{F\_27} & 0.1311 & 0.0925 & 0.0047 & 0.4105 & 0.9446 & -0.1943 \\
  \texttt{F\_28} & 0.1355 & 0.0961 & 0.0048 & 0.4280 & 0.9616 & -0.1545 \\
  \texttt{F\_29} & 0.1402 & 0.0998 & 0.0050 & 0.4463 & 0.9787 & -0.1142 \\
  \texttt{F\_30} & 0.1449 & 0.1037 & 0.0051 & 0.4638 & 0.9958 & -0.0723 \\
  \texttt{F\_31} & 0.1499 & 0.1077 & 0.0053 & 0.4835 & 1.0131 & -0.0290 \\
  \texttt{F\_32} & 0.1549 & 0.1119 & 0.0054 & 0.5038 & 1.0306 & 0.0156 \\
  \texttt{F\_33} & 0.1602 & 0.1162 & 0.0056 & 0.5252 & 1.0481 & 0.0611 \\
  \texttt{F\_34} & 0.1656 & 0.1206 & 0.0057 & 0.5476 & 1.0658 & 0.1077 \\
  \texttt{F\_35} & 0.1712 & 0.1253 & 0.0062 & 0.5709 & 1.0836 & 0.1557 \\
  \texttt{F\_36} & 0.1769 & 0.1300 & 0.0066 & 0.5953 & 1.1015 & 0.2052 \\
  \texttt{F\_37} & 0.1829 & 0.1350 & 0.0072 & 0.6206 & 1.1195 & 0.2564 \\
  \texttt{F\_38} & 0.1890 & 0.1401 & 0.0079 & 0.6453 & 1.1376 & 0.3084 \\
  \texttt{F\_39} & 0.1953 & 0.1453 & 0.0082 & 0.6728 & 1.1559 & 0.3619 \\
  \texttt{F\_40} & 0.2019 & 0.1508 & 0.0085 & 0.7014 & 1.1742 & 0.4168 \\
  \texttt{F\_41} & 0.2086 & 0.1564 & 0.0087 & 0.7295 & 1.1927 & 0.4734 \\
  \texttt{F\_42} & 0.2155 & 0.1622 & 0.0090 & 0.7606 & 1.2112 & 0.5314 \\
  \texttt{F\_43} & 0.2226 & 0.1682 & 0.0095 & 0.7930 & 1.2299 & 0.5906 \\
  \texttt{F\_44} & 0.2300 & 0.1744 & 0.0100 & 0.8267 & 1.2486 & 0.6517 \\
  \texttt{F\_45} & 0.2376 & 0.1808 & 0.0105 & 0.8620 & 1.2675 & 0.7143 \\
  \texttt{F\_46} & 0.2454 & 0.1874 & 0.0110 & 0.8981 & 1.2864 & 0.7786 \\
  \texttt{F\_47} & 0.2534 & 0.1942 & 0.0119 & 0.9358 & 1.3055 & 0.8443 \\
  \texttt{F\_48} & 0.2617 & 0.2012 & 0.0122 & 0.9753 & 1.3246 & 0.9114 \\
  \texttt{F\_49} & 0.2703 & 0.2085 & 0.0129 & 1.0169 & 1.3437 & 0.9803 \\
  \texttt{F\_50} & 0.2791 & 0.2159 & 0.0138 & 1.0602 & 1.3629 & 1.0512 \\
  \texttt{F\_51} & 0.2881 & 0.2236 & 0.0144 & 1.1054 & 1.3822 & 1.1235 \\
  \texttt{F\_52} & 0.2975 & 0.2314 & 0.0149 & 1.1525 & 1.4014 & 1.1972 \\
  \texttt{F\_53} & 0.3071 & 0.2395 & 0.0155 & 1.2001 & 1.4207 & 1.2722 \\
  \texttt{F\_54} & 0.3170 & 0.2479 & 0.0162 & 1.2512 & 1.4399 & 1.3490 \\
  \texttt{F\_55} & 0.3272 & 0.2564 & 0.0168 & 1.3045 & 1.4592 & 1.4273 \\
  \texttt{F\_56} & 0.3376 & 0.2652 & 0.0182 & 1.3601 & 1.4783 & 1.5072 \\
  \texttt{F\_57} & 0.3484 & 0.2743 & 0.0200 & 1.4165 & 1.4973 & 1.5882 \\
  \texttt{F\_58} & 0.3595 & 0.2835 & 0.0206 & 1.4769 & 1.5162 & 1.6703 \\
  \texttt{F\_59} & 0.3710 & 0.2930 & 0.0212 & 1.5398 & 1.5350 & 1.7536 \\
  \texttt{F\_60} & 0.3827 & 0.3027 & 0.0238 & 1.6010 & 1.5535 & 1.8378 \\
  \texttt{F\_61} & 0.3948 & 0.3127 & 0.0245 & 1.6678 & 1.5718 & 1.9228 \\
  \texttt{F\_62} & 0.4072 & 0.3229 & 0.0252 & 1.7152 & 1.5899 & 2.0081 \\
  \texttt{F\_63} & 0.4200 & 0.3334 & 0.0261 & 1.7883 & 1.6075 & 2.0935 \\
  \texttt{F\_64} & 0.4332 & 0.3441 & 0.0277 & 1.8633 & 1.6247 & 2.1790 \\
  \texttt{F\_65} & 0.4468 & 0.3550 & 0.0300 & 1.9427 & 1.6415 & 2.2642 \\
  \texttt{F\_66} & 0.4607 & 0.3662 & 0.0322 & 2.0255 & 1.6578 & 2.3486 \\
  \texttt{F\_67} & 0.4750 & 0.3776 & 0.0331 & 2.1006 & 1.6735 & 2.4317 \\
  \texttt{F\_68} & 0.4898 & 0.3893 & 0.0346 & 2.1899 & 1.6886 & 2.5130 \\
  \texttt{F\_69} & 0.5050 & 0.4012 & 0.0370 & 2.2832 & 1.7030 & 2.5926 \\
  \texttt{F\_70} & 0.5206 & 0.4134 & 0.0400 & 2.3752 & 1.7166 & 2.6698 \\
  \texttt{F\_71} & 0.5367 & 0.4258 & 0.0418 & 2.4764 & 1.7294 & 2.7442 \\
  \texttt{F\_72} & 0.5532 & 0.4384 & 0.0430 & 2.5819 & 1.7413 & 2.8151 \\
  \texttt{F\_73} & 0.5703 & 0.4514 & 0.0442 & 2.6877 & 1.7524 & 2.8819 \\
  \texttt{F\_74} & 0.5878 & 0.4646 & 0.0454 & 2.8022 & 1.7626 & 2.9445 \\
  \texttt{F\_75} & 0.6059 & 0.4781 & 0.0468 & 2.9195 & 1.7719 & 3.0028 \\
  \texttt{F\_76} & 0.6245 & 0.4919 & 0.0472 & 3.0342 & 1.7802 & 3.0563 \\
  \texttt{F\_77} & 0.6436 & 0.5059 & 0.0472 & 3.1447 & 1.7877 & 3.1050 \\
  \texttt{F\_78} & 0.6633 & 0.5204 & 0.0472 & 3.2787 & 1.7943 & 3.1489 \\
  \texttt{F\_79} & 0.6837 & 0.5351 & 0.0472 & 3.4168 & 1.8001 & 3.1876 \\
  \texttt{F\_80} & 0.7046 & 0.5502 & 0.0472 & 3.5624 & 1.8052 & 3.2218 \\
  \texttt{F\_81} & 0.7262 & 0.5656 & 0.0472 & 3.7139 & 1.8096 & 3.2517 \\
  \texttt{F\_82} & 0.7485 & 0.5814 & 0.0472 & 3.8721 & 1.8134 & 3.2776 \\
  \texttt{F\_83} & 0.7714 & 0.5977 & 0.0472 & 4.0337 & 1.8166 & 3.3000 \\
  \texttt{F\_84} & 0.7951 & 0.6143 & 0.0472 & 4.1839 & 1.8195 & 3.3192 \\
  \texttt{F\_85} & 0.8195 & 0.6314 & 0.0472 & 4.3235 & 1.8219 & 3.3355 \\
  \texttt{F\_86} & 0.8446 & 0.6489 & 0.0472 & 4.5077 & 1.8241 & 3.3498 \\
  \texttt{F\_87} & 0.8706 & 0.6669 & 0.0472 & 4.6765 & 1.8261 & 3.3624 \\
  \texttt{F\_88} & 0.8973 & 0.6854 & 0.0472 & 4.8738 & 1.8278 & 3.3738 \\
  \texttt{F\_89} & 0.9249 & 0.7044 & 0.0472 & 5.0484 & 1.8295 & 3.3844 \\
  \texttt{F\_90} & 0.9534 & 0.7239 & 0.0472 & 5.2593 & 1.8311 & 3.3945 \\
  \texttt{F\_91} & 0.9827 & 0.7439 & 0.0472 & 5.4833 & 1.8326 & 3.4041 \\
  \texttt{F\_92} & 1.0130 & 0.7645 & 0.0472 & 5.7169 & 1.8341 & 3.4140 \\
  \texttt{F\_93} & 1.0442 & 0.7856 & 0.0472 & 5.9403 & 1.8356 & 3.4243 \\
  \texttt{F\_94} & 1.0764 & 0.8074 & 0.0472 & 6.1934 & 1.8372 & 3.4352 \\
  \texttt{F\_95} & 1.1096 & 0.8297 & 0.0472 & 6.4005 & 1.8386 & 3.4467 \\
  \texttt{F\_96} & 1.1438 & 0.8526 & 0.0472 & 6.6732 & 1.8401 & 3.4586 \\
  \texttt{F\_97} & 1.1791 & 0.8761 & 0.0472 & 6.9564 & 1.8415 & 3.4707 \\
  \texttt{F\_98} & 1.2156 & 0.9003 & 0.0472 & 7.2131 & 1.8428 & 3.4831 \\
  \texttt{F\_99} & 1.2531 & 0.9252 & 0.0472 & 7.3848 & 1.8439 & 3.4954 \\
  \texttt{F\_100} & 1.2919 & 0.9507 & 0.0472 & 7.5600 & 1.8449 & 3.5075 \\
  \texttt{F\_101} & 1.3318 & 0.9769 & 0.0472 & 7.7427 & 1.8457 & 3.5189 \\
  \texttt{F\_102} & 1.3730 & 1.0038 & 0.0472 & 7.9332 & 1.8463 & 3.5297 \\
  \texttt{F\_103} & 1.4155 & 1.0314 & 0.0472 & 8.1318 & 1.8467 & 3.5393 \\
  \texttt{F\_104} & 1.4594 & 1.0598 & 0.0472 & 8.3388 & 1.8468 & 3.5479 \\
  \texttt{F\_105} & 1.5046 & 1.0889 & 0.0472 & 8.5547 & 1.8466 & 3.5554 \\
  \texttt{F\_106} & 1.5512 & 1.1189 & 0.0472 & 8.7797 & 1.8462 & 3.5618 \\
  \texttt{F\_107} & 1.5993 & 1.1497 & 0.0472 & 9.0144 & 1.8455 & 3.5670 \\
  \texttt{F\_108} & 1.6489 & 1.1813 & 0.0472 & 9.2590 & 1.8445 & 3.5711 \\
  \texttt{F\_109} & 1.7001 & 1.2138 & 0.0472 & 9.5141 & 1.8433 & 3.5738 \\
  \texttt{F\_110} & 1.7528 & 1.2473 & 0.0472 & 9.7800 & 1.8418 & 3.5753 \\
  \texttt{F\_111} & 1.8072 & 1.2817 & 0.0472 & 10.0573 & 1.8400 & 3.5758 \\
  \texttt{F\_112} & 1.8633 & 1.3170 & 0.0472 & 10.3463 & 1.8379 & 3.5750 \\
  \texttt{F\_113} & 1.9211 & 1.3534 & 0.0472 & 10.6477 & 1.8355 & 3.5731 \\
  \texttt{F\_114} & 1.9808 & 1.3908 & 0.0472 & 10.9619 & 1.8328 & 3.5700 \\
  \texttt{F\_115} & 2.0423 & 1.4292 & 0.0472 & 11.2907 & 1.8297 & 3.5656 \\
  \texttt{F\_116} & 2.1057 & 1.4688 & 0.0472 & 11.6343 & 1.8263 & 3.5600 \\
  \texttt{F\_117} & 2.1711 & 1.5096 & 0.0472 & 11.9925 & 1.8226 & 3.5533 \\
  \texttt{F\_118} & 2.2385 & 1.5516 & 0.0472 & 12.3659 & 1.8185 & 3.5453 \\
  \texttt{F\_119} & 2.3079 & 1.5948 & 0.0472 & 12.7553 & 1.8140 & 3.5363 \\
  \texttt{F\_120} & 2.3796 & 1.6393 & 0.0472 & 13.1612 & 1.8091 & 3.5259 \\
  \texttt{F\_121} & 2.4534 & 1.6851 & 0.0472 & 13.5845 & 1.8038 & 3.5142 \\
  \texttt{F\_122} & 2.5294 & 1.7323 & 0.0472 & 14.0257 & 1.7981 & 3.5012 \\
  \texttt{F\_123} & 2.6078 & 1.7810 & 0.0472 & 14.4608 & 1.7919 & 3.4868 \\
  \texttt{F\_124} & 2.6886 & 1.8312 & 0.0472 & 14.8291 & 1.7852 & 3.4709 \\
  \texttt{F\_125} & 2.7718 & 1.8829 & 0.0472 & 15.2130 & 1.7780 & 3.4535 \\
  \texttt{F\_126} & 2.8575 & 1.9363 & 0.0472 & 15.6134 & 1.7703 & 3.4343 \\
  \texttt{F\_127} & 2.9459 & 1.9913 & 0.0472 & 16.0307 & 1.7621 & 3.4135 \\
  \texttt{N\_asperities} & 2917.76 & 3822.78 & 30.00 & 12000.00 & 1.2250 & 0.0901 \\
\end{longtable}
} 

\begin{table}[htbp]
  \centering
  \caption{Descriptive statistics for the target GMM parameters.}
  \label{tab:desc_stats_targets}
  \begin{tabular}{@{}lrrrrrr@{}}
    \toprule
    Parameter & Mean & Std Dev & Min & Max & Skewness & Kurtosis \\
    \midrule
    $w_1$ & 0.5000 & 0.2886 & 0.0000 & 1.0000 & 0.0001 & -1.2005 \\
    $w_2$ & 0.5000 & 0.2886 & 0.0000 & 1.0000 & 0.0001 & -1.2005 \\
    $w_3$ & 0.5000 & 0.2886 & 0.0000 & 1.0000 & 0.0001 & -1.2005 \\
    $\mu_{h,1}$ & 150.0000 & 57.7339 & 50.0000 & 250.0000 & -0.0003 & -1.1938 \\
    $\mu_{R,1}$ & 275.0011 & 129.9014 & 50.0000 & 500.0000 & -0.0004 & -1.2034 \\
    $\mu_{h,2}$ & 150.0000 & 57.7339 & 50.0000 & 250.0000 & -0.0003 & -1.1938 \\
    $\mu_{R,2}$ & 275.0012 & 129.9013 & 50.0000 & 500.0000 & -0.0004 & -1.2034 \\
    $\mu_{h,3}$ & 150.0000 & 57.7339 & 50.0000 & 250.0000 & -0.0003 & -1.1938 \\
    $\mu_{R,3}$ & 275.0012 & 129.9014 & 50.0000 & 499.9999 & -0.0004 & -1.2034 \\
    $\mu_{h,4}$ & 150.0000 & 57.7339 & 50.0000 & 250.0000 & -0.0003 & -1.1938 \\
    $\mu_{R,4}$ & 275.0011 & 129.9014 & 50.0000 & 500.0000 & -0.0004 & -1.2034 \\
    $\sigma_{h,1}$ & 45.0000 & 20.2077 & 10.0000 & 80.0000 & 0.0002 & -1.2107 \\
    $\sigma_{R,1}$ & 55.0000 & 25.9817 & 10.0000 & 100.0000 & -0.0009 & -1.2016 \\
    $\sigma_{h,2}$ & 45.0000 & 20.2077 & 10.0000 & 80.0000 & 0.0002 & -1.2107 \\
    $\sigma_{R,2}$ & 55.0000 & 25.9818 & 10.0000 & 100.0000 & -0.0010 & -1.2016 \\
    $\sigma_{h,3}$ & 45.0000 & 20.2077 & 10.0000 & 80.0000 & 0.0002 & -1.2107 \\
    $\sigma_{R,3}$ & 55.0000 & 25.9817 & 10.0000 & 100.0000 & -0.0009 & -1.2016 \\
    $\sigma_{h,4}$ & 45.0000 & 20.2077 & 10.0000 & 80.0000 & 0.0002 & -1.2107 \\
    $\sigma_{R,4}$ & 55.0000 & 25.9817 & 10.0000 & 100.0000 & -0.0009 & -1.2016 \\
    $\rho_1$ & 0.0000 & 0.5196 & -0.9000 & 0.9000 & 0.0000 & -1.2001 \\
    $\rho_2$ & 0.0000 & 0.5196 & -0.9000 & 0.9000 & 0.0000 & -1.2001 \\
    $\rho_3$ & 0.0000 & 0.5196 & -0.9000 & 0.9000 & 0.0000 & -1.2001 \\
    $\rho_4$ & 0.0000 & 0.5196 & -0.9000 & 0.9000 & 0.0000 & -1.2001 \\
    \bottomrule
  \end{tabular}
\end{table}

\begin{figure}[h!]
    \centering
    \includegraphics[width=0.65\linewidth]{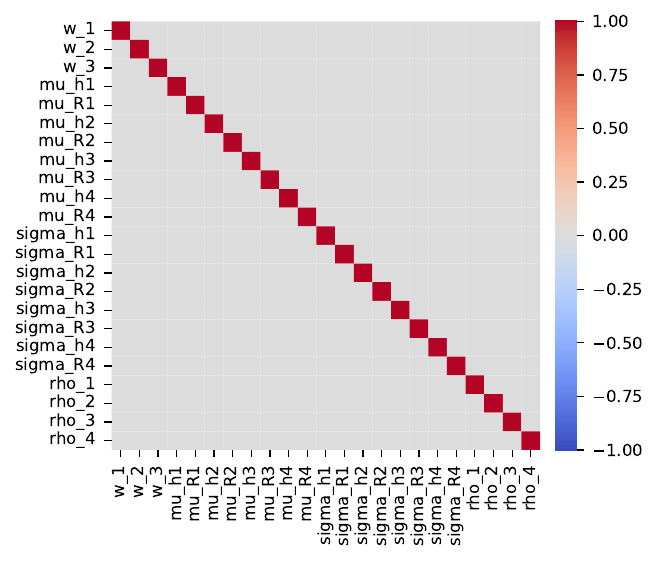}
    \caption{Pearson correlation matrix of the 23 target GMM parameters. The axes represent the individual parameters, grouped by type (weights, means, standard deviations, and correlations).}
    \label{fig:corr_yy}
\end{figure}

\begin{figure}[h!]
    \centering
    \includegraphics[width=0.72\linewidth]{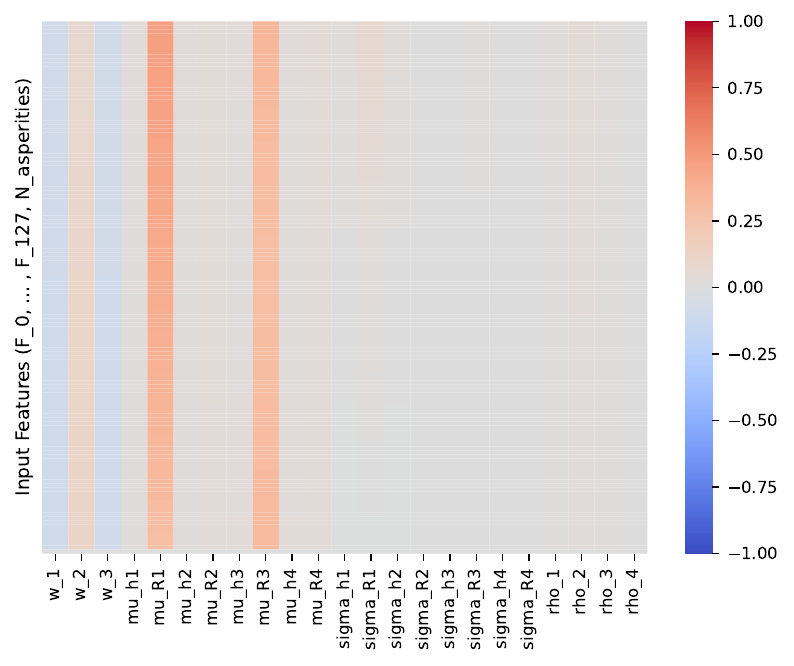}
    \caption{Pearson correlation matrix between the 129 input features (y-axis) and the 23 target GMM parameters (x-axis). The input features comprise the 128 points of the discretized friction law and the asperity count ($N_{\text{asperities}}$).}
    \label{fig:corr_xy}
\end{figure}

To complete the statistical overview, we analyzed the Pearson correlation coefficients between variables. The resulting heatmaps are shown in Figures~\ref{fig:corr_yy} and \ref{fig:corr_xy}. The Target-vs-Target correlation matrix (Figure~\ref{fig:corr_yy}) shows near-zero off-diagonal correlations, consistent with a quasi-linearly independent parameter space generated by the Sobol sampling strategy. The Feature-vs-Target matrix (Figure~\ref{fig:corr_xy}) indicates a positive correlation between the friction law values \(F_i\) and the mean asperity radii~\(\mu_{R,k}\).

\subsection{Friction Law Examples}

\begin{figure}[p!] 
    \centering
    \vspace{-2.5em}
    \begin{subfigure}{\textwidth}
        \centering
        \includegraphics[width=0.7\linewidth]{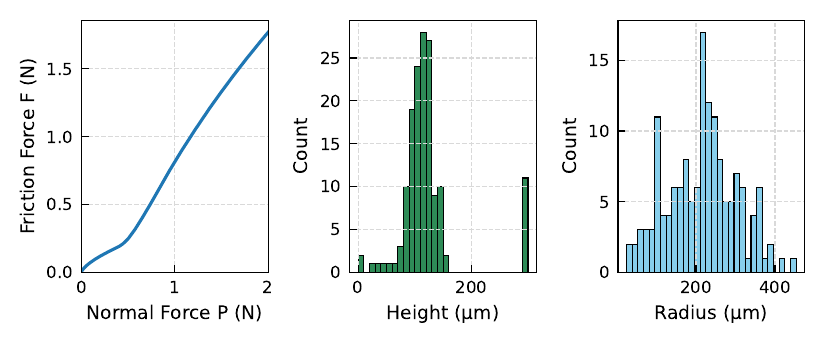}
        \caption{A piecewise non-linear friction law with \(N = 150\) asperities.}
        \label{fig:sample-piecewise}
    \end{subfigure}
    \vfill 
    
    \begin{subfigure}{\textwidth}
        \centering
        \includegraphics[width=0.7\linewidth]{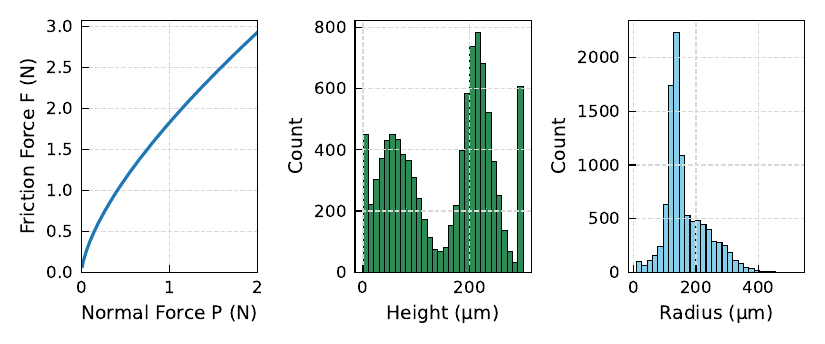}
        \caption{A concave non-linear friction law with \(N = 10,000\) asperities.}
        \label{fig:sample-concave}
    \end{subfigure}
    \vfill

    \begin{subfigure}{\textwidth}
        \centering
        \includegraphics[width=0.7\linewidth]{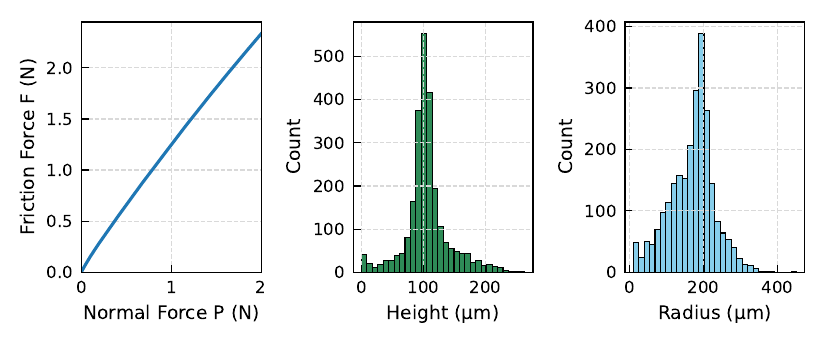}
        \caption{A quasi-linear friction law with \(N = 2,500\) asperities.}
        \label{fig:sample-linear}
    \end{subfigure}
    \vfill

    \begin{subfigure}{\textwidth}
        \centering
        \includegraphics[width=0.7\linewidth]{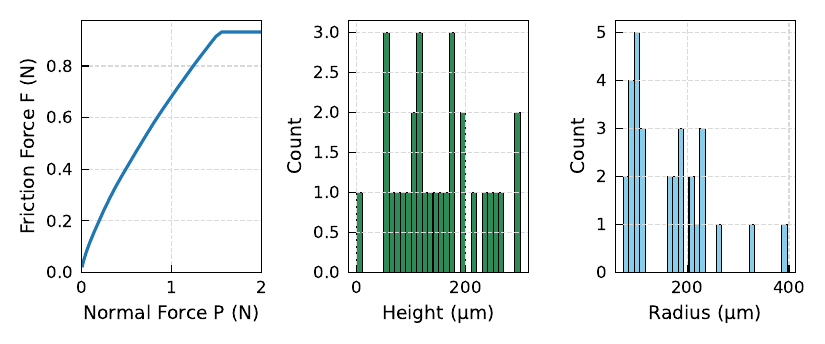}
        \caption{A saturating concave non-linear friction law with \(N = 30\) asperities.}
        \label{fig:sample-saturating}
    \end{subfigure}
    \vfill

    \begin{subfigure}{\textwidth}
        \centering
        \includegraphics[width=0.7\linewidth]{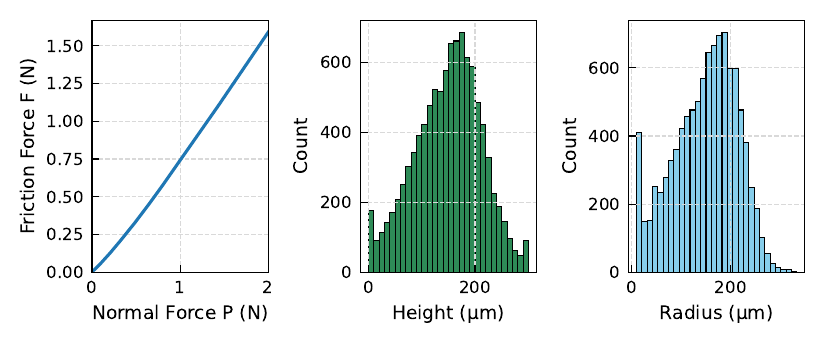}
        \caption{A convex non-linear friction law with \(N = 10,000\) asperities.}
        \label{fig:sample-convex}
    \end{subfigure}

    \caption{Examples of the various possible friction laws and their corresponding asperity height and radius distributions. Each row corresponds to a single, specific data sample.}
    \label{fig:specific-samples-appendix}
\end{figure}

The synthetic dataset encompasses a wide spectrum of physically valid frictional behaviors. Figure~\ref{fig:specific-samples-appendix} provides representative examples from this dataset, illustrating the direct correspondence between 

macroscopic friction laws and their underlying microscopic surface topographies. Each row corresponds to a single data sample, displaying the discretized friction force curve \(\bm{F}(P)\) alongside the histograms of the asperity height and radius distributions that generate it. While the selected samples showcase the dataset's richness by covering a range of behaviors from quasi-linear to piecewise non-linear friction laws, it is important to note that this diversity is not uniformly distributed despite the uniform sampling of the GMM parameters. The standard concave non-linear law (e.g., Figure~\ref{fig:sample-concave}) represents the dominant mode and may explain the CVAE's tendency to output this common friction law shape. One may also note, for instance, how the bimodal height distribution generates a piecewise non-linear law, whereas a distribution resembling a truncated exponential yields a quasi-linear response, a behavior consistent with the classic GW model.

\clearpage
\section{Model Architectures}
\label{appendix:model}

This section details the architectures of the primary models evaluated in this study: the Conditional Variational Autoencoder (CVAE), the unconditional Variational Autoencoder (VAE), and two baselines, a Multi-Layer Perceptron (MLP) and an XGBoost model. The final CVAE hyperparameters were determined through an extensive optimization study using the Optuna framework. For a direct comparison of regression versus generative capabilities, the MLP baseline was architecturally matched to the CVAE's decoder. All neural network models were implemented in PyTorch, and the XGBoost model was trained on a GPU using the \texttt{cupy} backend (cf. Appendix~\ref{appendix:setup}).

\subsection{CVAE Architecture}

The CVAE forms the core of our generative framework. Its architecture and key training hyperparameters are specified in Table~\ref{tab:cvae_architecture}.

\begin{table}[h!]
\centering
\caption{Architectural and training specifications for the best-performing CVAE model, as determined by the Optuna hyperparameter search (Trial \#303).}
\label{tab:cvae_architecture}
\begin{tabular}{lll}
\toprule
\textbf{Component} & \textbf{Parameter} & \textbf{Specification} \\
\midrule
\multirow{5}{*}{\textbf{Encoder}} & Input Dimensions & 152 (129 condition + 23 GMM parameters) \\
 & Hidden Layers & 3 \\
 & \quad \textit{Layer 1} & 1915 units (Dropout: 0.163) \\
 & \quad \textit{Layer 2} & 1723 units (Dropout: 0.080) \\
 & \quad \textit{Layer 3} & 767 units (Dropout: 0.090) \\
\midrule
\textbf{Latent Space} & Dimensions & 56 \\
\midrule
\multirow{5}{*}{\textbf{Decoder}} & Input Dimensions & 185 (129 condition + 56 latent parameters) \\
 & Hidden Layers & 3 \\
 & \quad \textit{Layer 1} & 347 units (Dropout: 0.024) \\
 & \quad \textit{Layer 2} & 308 units (Dropout: 0.073) \\
 & \quad \textit{Layer 3} & 328 units (Dropout: 0.123) \\
\midrule
\multirow{3}{*}{\textbf{Shared}} & Internal Block & Linear $\rightarrow$ BatchNorm1d $\rightarrow$ PReLU $\rightarrow$ Dropout \\
 & Output Activation & \texttt{tanh} \\
 & Weight Initialization & Kaiming Normal (for Linear layers) \\
\midrule
\multirow{5}{*}{\textbf{Training}} & Optimizer & AdamW \\
 & Batch Size & 8192 \\
 & Max Learning Rate & $1.98 \times 10^{-4}$ (with OneCycleLR) \\
 & Weight Decay & $1.10 \times 10^{-6}$ \\
 & KL Beta ($\beta_{final}$) & $1.06 \times 10^{-5}$ \\
\bottomrule
\end{tabular}
\end{table}

\subsection{Derived and Baseline Model Architectures}

The remaining models were either derived from the CVAE architecture or were standard ML baselines tuned for this task.

\textbf{Unconditional VAE.}~The VAE used for the optimization-based benchmark shares the exact same architecture as the CVAE detailed in Table~\ref{tab:cvae_architecture}. The sole modification is the omission of the conditional inputs. This reduces the encoder input to 23 dimensions (GMM parameters only) and the decoder input to 56 dimensions (the latent vector only).

\textbf{Multi-Layer Perceptron.}~The MLP baseline directly mirrors the CVAE's decoder architecture to provide a fair, non-generative comparison. It accepts the 129-dimensional condition vector as input and processes it through the same three hidden layers (347, 308, and 328 units), dropout rates, and internal block structure as the CVAE decoder. It was trained using the same optimizer, batch size, and learning rate schedule.

\textbf{XGBoost.}~The XGBoost baseline consists of 23 independent gradient-boosted decision tree models, one for each dimension of the target GMM parameter vector. The models were trained iteratively in chunks to handle the large dataset. The key hyperparameters, identical for all 23 models, are listed in Table~\ref{tab:xgboost_params}.

\begin{table}[h!]
\centering
\caption{Key hyperparameters for the XGBoost baseline models.}
\label{tab:xgboost_params}
\begin{tabular}{llc}
\toprule
\textbf{Hyperparameter} & \textbf{Description} & \textbf{Value} \\
\midrule
\texttt{learning\_rate} & Step-size shrinkage ($\eta$) & 0.05 \\
\texttt{max\_depth} & Maximum tree depth & 8 \\
\texttt{subsample} & Subsample ratio of training instances & 0.8 \\
\texttt{colsample\_bytree} & Subsample ratio of columns for each tree & 0.8 \\
\texttt{gamma} & Minimum loss reduction for split & 0.1 \\
\texttt{lambda} & L2 regularization term & 1 \\
\texttt{alpha} & L1 regularization term & 0.1 \\
\texttt{device} & Hardware backend for training & \texttt{cuda} \\
\bottomrule
\end{tabular}
\end{table}

\subsection{Post-processing for Physical Validity}
\label{appendix:clamping}

The generative nature of the CVAE, combined with a standard \texttt{tanh} output activation, ensures that individual parameter outputs are bounded within their scaled [-1, 1] range. However, this architecture does not inherently enforce all physical constraints of the GMM parameter space after the outputs are unscaled. For instance, the sum of the first three mixture weights may exceed 1. To address this, a two-step post-processing function is applied to the raw, unscaled model outputs to guarantee their physical validity. The post-processing function operates as follows:

\textbf{Clamping.} The first step involves a direct clamping operation. Each of the 23 generated GMM parameters is individually clipped using its corresponding minimum and maximum values to ensure it falls within the physical bounds defined during dataset generation (see Table~\ref{tab:gmm_bounds}).

\textbf{Weight Normalization.} The second step addresses the collective constraint on the GMM mixture weights. After the initial clamping, the procedure checks if the sum of the first three generated weights, \(\sum_{k=1}^{3} w_k\), exceeds 1. In cases where it does, these three weights are proportionally scaled by dividing each one by their sum. This normalization ensures that the corrected weights (\(w'_k = w_k / \sum_{i=1}^{3} w_i\)) now sum exactly to 1, while preserving their relative contributions. The fourth weight, \(w_4\), is then implicitly defined as \(1 - \sum_{k=1}^{3} w'_k\).

This two-step post-processing function guarantees that all of the model's outputs are converted into physically valid GMM parameter sets before being used in any downstream analysis or forward simulation.

\subsection{Hyperparameter Optimization}
\label{sec:app_appendix_hyperparams}

This section provides the complete details of the CVAE hyperparameter optimization.

\begin{table*}[ht]
  \caption{Hyperparameter and methodological configuration for the CVAE optimization.}
  \label{tab:cvae-optuna-methodology-full} 
  \centering
  \sisetup{round-mode=places, round-precision=2, group-separator={,}}
  \begin{tabular}{ll}
    \toprule
    Category & Description / Sampling Range \\
    \midrule
    \multicolumn{2}{l}{\textit{Optuna Study Configuration}} \\
    Objective Metric & Minimize Validation Loss = SmoothL1(x, \(\hat{x}\)) + \(\beta_{KL}\)\,D\(_{KL}\) \\
    Total Trials & 331 \\
    Optuna Sampler & Tree-structured Parzen Estimator Sampler (Seed: 12) \\
    Pruning Strategy & MedianPruner (Startup: 5, Warmup: 80,000 steps, Interval: 1) \\
    \midrule
    \multicolumn{2}{l}{\textit{Data Configuration}} \\
    Dataset Size & 200,278,016 samples \\
    Source & Iterable dataset constructed from pre-processed \texttt{.pt} shards \\
    File-level Split & 70\% Train / 15\% Validation / 15\% Test \\
    Split Seed & 42 (applied at the shard level) \\
    \midrule
    \multicolumn{2}{l}{\textit{Model \& Training Details}} \\
    Model Variant & Conditional Variational Autoencoder (CVAE) \\
    Steps per Trial & 40,000 \\
    Optimizer & AdamW ($\beta_1 = 0.9$, $\beta_2 = 0.999$) \\
    Scheduler & OneCycleLR \\
    \midrule
    \multicolumn{2}{l}{\textit{Training Hyperparameters (Searched)}} \\
    Batch Size & Categorical \{128, 256, 512, 1024, 2048, 4096, 8192, 16384\} \\
    Learning Rate (max\_lr) & Log-float \([10^{-5}, 2 \times 10^{-4}]\) \\
    Weight Decay & Log-float \([10^{-7}, 10^{-3}]\) \\
    KL Annealing Beta (\(\beta_{KL}\)) & Log-float \([10^{-5}, 10^{-2}]\) (with a Linear Warmup on 20,000 Steps) \\
    \midrule
    \multicolumn{2}{l}{\textit{Architectural Hyperparameters (Searched)}} \\
    Latent Dimension & Integer \([24, 64]\) \\
    Encoder Hidden Layers & Integer \([2, 6]\) \\
    Encoder Units per HL & Log-integer \([64, 2048]\) \\
    Encoder Dropout per HL & Float \([0.0, 0.5]\) \\
    Decoder Hidden Layers & Integer \([2, 6]\) \\
    Decoder Units per HL & Log-integer \([64, 2048]\) \\
    Decoder Dropout per HL & Float \([0.0, 0.5]\) \\
    \bottomrule
  \end{tabular}
\end{table*}

\subsection{Ablation Studies}
\label{appendix:ablation}

We conducted ablation experiments to evaluate the impact of latent dimensionality, KL regularization weight, batch size, training duration, stochasticity, and conditioning on model performance (Appendix~\ref{appendix:ablation}, Tables~\ref{tab:ablation_latent_dim}–\ref{tab:ablation_r2}). Each study was performed under otherwise identical conditions to isolate the contribution of the tested variable. Note that the baseline metrics change as the model is retrained with each ablation.

The most pronounced effect arises from conditioning: removing the conditional input (unconditional VAE) reduces parameter-level sMAPE from 2.97\% to 1.70\%, which indicates that conditioning slightly degrades the reconstruction accuracy at the parameter level. Extreme reductions in latent dimensionality (e.g., 16) severely degrade performance (sMAPE > 20\%), whereas moderate increases (up to 128 dimensions) yield modest improvements. Higher KL weighting generally increases error, and very small batch sizes also deteriorate accuracy. Training duration and stochasticity variations produced comparatively minor effects. Adjusted \(R^2\) scores corroborate these trends, confirming that conditioning and latent size are the most influential factors for predictive accuracy.

\subsection{Latent Dimensionality}
\begin{table}[H]
\centering
\caption{Effect of latent dimensionality on validation sMAPE and loss.}
\label{tab:ablation_latent_dim}
\begin{tabular}{lcc}
\toprule
Trial Name & Val sMAPE (\%) & Val Loss \\
\midrule
baseline\_control & 2.9143 & 0.00646 \\
latent\_dim\_16    & 20.0951 & 0.53572 \\
latent\_dim\_32    & 4.3139 & 0.00907 \\
latent\_dim\_80    & 3.4748 & 0.00756 \\
latent\_dim\_128   & 2.7423 & 0.00865 \\
\bottomrule
\end{tabular}
\end{table}

\subsection{KL Regularization Weight}
\begin{table}[H]
\centering
\caption{Effect of KL regularization weight on validation sMAPE and loss.}
\begin{tabular}{lcc}
\toprule
Trial Name & Val sMAPE (\%) & Val Loss \\
\midrule
baseline\_control       & 2.4730 & 0.00623 \\
beta\_kl\_final\_0.0001 & 2.9505 & 0.01840 \\
beta\_kl\_final\_0.001  & 3.9574 & 0.07903 \\
beta\_kl\_final\_0.01   & 4.1737 & 0.40140 \\
\bottomrule
\end{tabular}
\end{table}

\subsection{Batch Size}
\begin{table}[H]
\centering
\caption{Effect of batch size on validation sMAPE and loss.}
\begin{tabular}{lcc}
\toprule
Trial Name & Val sMAPE (\%) & Val Loss \\
\midrule
baseline\_control & 3.0647 & 0.00842 \\
batch\_size\_2048  & 3.8574 & 0.00936 \\
batch\_size\_512   & 6.2008 & 0.02782 \\
\bottomrule
\end{tabular}
\end{table}

\subsection{Training Steps}
\begin{table}[H]
\centering
\caption{Effect of total training steps on validation sMAPE and loss.}
\begin{tabular}{lcc}
\toprule
Trial Name & Val sMAPE (\%) & Val Loss \\
\midrule
baseline\_control  & 3.7095 & 0.00806 \\
total\_steps\_30000 & 3.1012 & 0.00929 \\
total\_steps\_20000 & 5.0588 & 0.01505 \\
\bottomrule
\end{tabular}
\end{table}

\subsection{Stochasticity}
\begin{table}[H]
\centering
\caption{Effect of model stochasticity on validation sMAPE and loss.}
\begin{tabular}{lcc}
\toprule
Trial Name & Val sMAPE (\%) & Val Loss \\
\midrule
baseline\_control & 3.4423 & 0.00862 \\
model\_type\_CAE   & 3.2390 & 0.00518 \\
\bottomrule
\end{tabular}
\end{table}

\subsection{Conditioning}
\begin{table}[H]
\centering
\caption{Effect of conditioning on validation sMAPE and loss.}
\begin{tabular}{lcc}
\toprule
Trial Name & Val sMAPE (\%) & Val Loss \\
\midrule
baseline\_control & 3.2538 & 0.00791 \\
model\_type\_VAE  & 1.6973 & 0.00374 \\
\bottomrule
\end{tabular}
\end{table}

\subsection{Weighted Adjusted \(R^2\) Scores}
\begin{table}[H]
\centering
\caption{Weighted average Adjusted \(R^2\) for ablation experiments.}
\label{tab:ablation_r2}
\begin{tabular}{lcc}
\toprule
Study & Trial Name & Weighted Avg Adjusted \(R^2\) \\
\midrule
Latent Dimensionality & baseline\_control & 0.998721 \\
Latent Dimensionality & latent\_dim\_16  & 0.764154 \\
Latent Dimensionality & latent\_dim\_32  & 0.997348 \\
Latent Dimensionality & latent\_dim\_80  & 0.998669 \\
Latent Dimensionality & latent\_dim\_128 & 0.998446 \\
KL Weight             & baseline\_control & 0.998260 \\
KL Weight             & beta\_kl\_final\_0.0001 & 0.997972 \\
KL Weight             & beta\_kl\_final\_0.001  & 0.994748 \\
KL Weight             & beta\_kl\_final\_0.01   & 0.985558 \\
Batch Size            & baseline\_control & 0.998253 \\
Batch Size            & batch\_size\_2048 & 0.998230 \\
Batch Size            & batch\_size\_512  & 0.994182 \\
Training Steps        & baseline\_control & 0.998275 \\
Training Steps        & total\_steps\_20000 & 0.996157 \\
Training Steps        & total\_steps\_30000 & 0.998593 \\
Stochasticity         & baseline\_control & 0.997766 \\
Stochasticity         & model\_type\_CAE  & 0.997838 \\
Conditioning          & baseline\_control & 0.997542 \\
Conditioning          & model\_type\_VAE  & 0.999727 \\
\bottomrule
\end{tabular}
\end{table}

\clearpage
\section{Additional Physical Results}
\subsection{Distributional Analysis of CVAE Outputs}
\label{appendix:distrib}
\begin{figure}[h!] 
    \centering
    \includegraphics[width=0.8\linewidth]{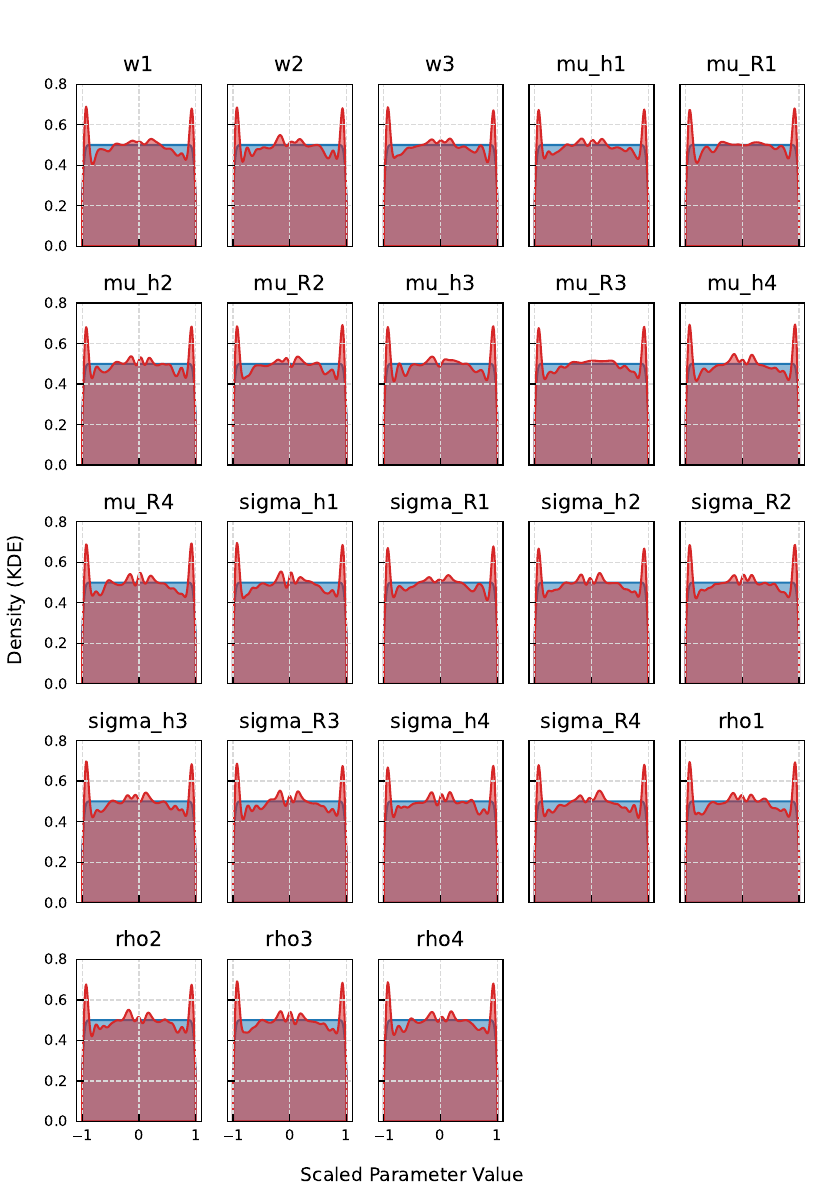}
    
    \caption{
        Comparison of the distributions for each of the 23 learned GMM parameters in the scaled space \([-1, 1]\). For each of the 23 GMM parameters, the Kernel Density Estimate (KDE) of the true data distribution (blue) is compared against the distribution generated by the CVAE (red). The close overlap across all dimensions indicates that the model has successfully learned the target manifold. The slight U-shaped bias in the CVAE's output, with peaks near the boundaries, is a characteristic artifact of the \texttt{tanh} output activation function.
    }
    \label{fig:dist-coverage} 
\end{figure}

Figure~\ref{fig:dist-coverage} shows the comparison of the distribution of all GMM parameters obtained with Kernel Density Estimate for the true data distribution and with the output of the CVAE in the scaled [-1, 1] space. We can see a satisfactory overlap of the two distributions, despite a slight U-shaped bias in the CVAE’s output due to the tanh output activation function. This, along with a Wasserstein distance close to zero, confirms that the CVAE is able to succesfully learn the target distributions.

\subsection{Predictive Uncertainty Quantification}
\label{appendix:uncertainty}

\begin{figure}[h!]
    \centering
    \includegraphics[width = 0.90\textwidth]{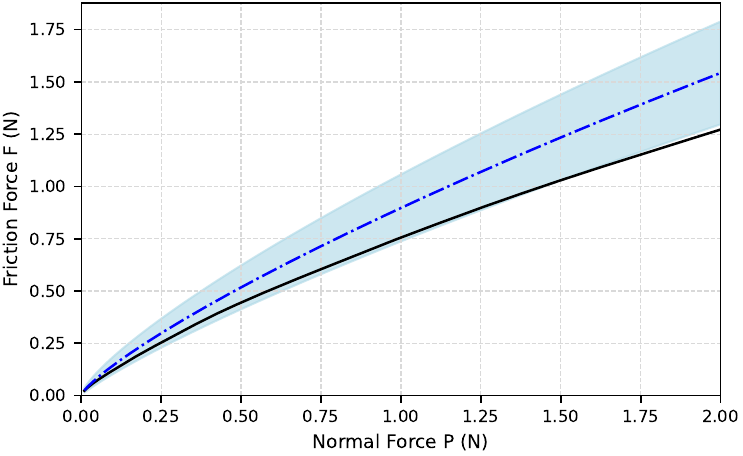}
    \caption{Illustration of the CVAE's probabilistic prediction for a single, unseen test sample. The solid black line represents the ground-truth friction law. The dash-dotted blue line is the mean predicted friction law, generated by averaging the output of 100,000 inferences from distinct latent space samples. The light blue shaded area represents the model's predictive uncertainty, corresponding to ±1 standard deviation around the mean. This envelope captures the diversity of valid solutions proposed by the generative model, with its deviation from the ground truth indicating the model's predictive bias for this specific case. Examples of the diverse surface topographies that can produce such friction laws are provided in Figure~\ref{fig:topo}.}
    \label{fig:cvae-uncertainty}
\end{figure}

\clearpage
\subsection{Sensitivity Analysis}
\label{appendix:sensitivity}

\begin{figure}[htbp]
    \centering
    \includegraphics[width=0.90\linewidth]{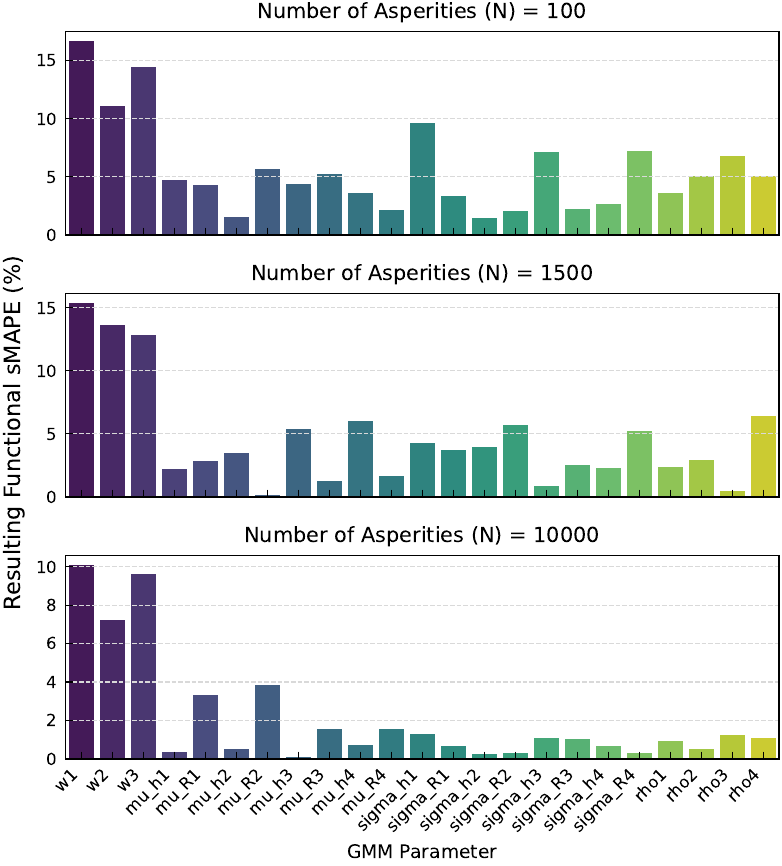}
    \caption{
            A sensitivity analysis quantifying the effect of GMM parameter perturbations on the simulated friction law.  Each panel corresponds to a simulation run with a different number of asperities: \(N=100\), \(N=1,500\), and \(N=10,00\), respectively. The height of each bar represents the functional sMAPE, measuring the deviation from a baseline friction curve when the corresponding GMM parameter on the x-axis is perturbed by 5\%. This analysis visually confirms the \emph{averaging} effect discussed in the main text: high-asperity systems are less sensitive to perturbations in individual shape parameters (e.g., mu\_h1, sigma\_R2) but remain sensitive to changes in mixture weights (w1, w2), which control the overall composition of the surface.}
    \label{fig:sensitivity_analysis}
\end{figure}

\clearpage
\section{Convergence}
\label{appendix:convergence}

\begin{figure}[h!]
    \centering
    \includegraphics[width=0.725\linewidth]{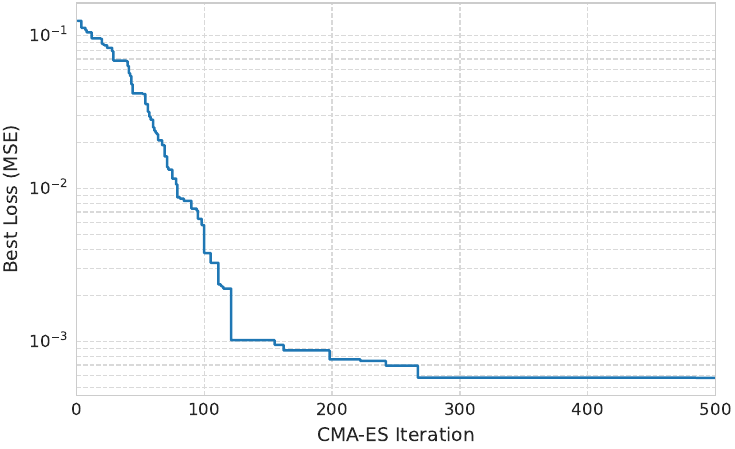}
    \caption{
        Convergence of the best-performing optimization run (Run 66). 
        The plot shows the best Mean Squared Error (MSE) loss found at each iteration of the CMA-ES algorithm. 
        The logarithmic scale on the y-axis highlights the rapid initial improvement and subsequent fine-tuning.
    }
    \label{fig:vae-convergence}
\end{figure}

\begin{figure}[h!]
    \centering
    \includegraphics[width=0.725\linewidth]{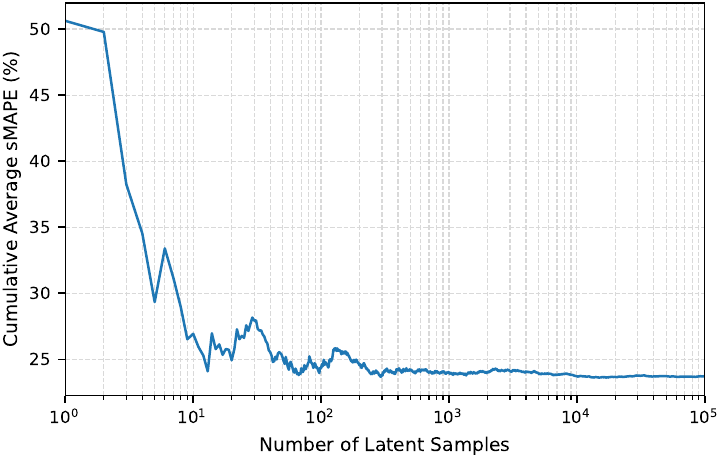}
    \caption{
        Convergence of the functional sMAPE for the CVAE model on a single test sample. The plot shows the cumulative average sMAPE as the number of stochastic latent samples increases. The logarithmic scale on the x-axis highlights the rapid decrease in average error with the initial samples, followed by a stable convergence to the final mean error value after approximately 10,000 samples.
    }
    \label{fig:cvae-convergence}
\end{figure}

\begin{figure}[p!] 
    \centering
    \vspace{-2em}
    \includegraphics{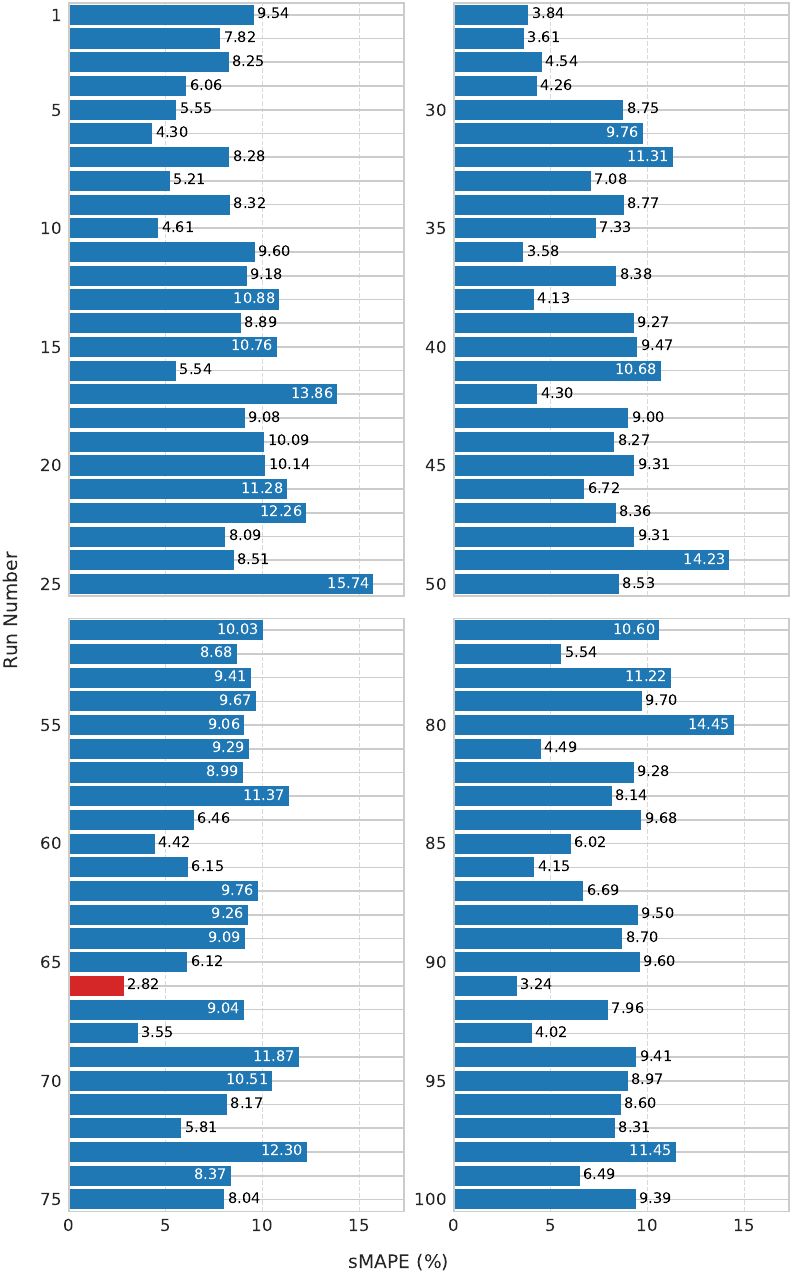}
    
    \caption{
        Symmetric Mean Absolute Percentage Error (sMAPE) for each of the 100 optimization runs, arranged in a 2x2 grid for clarity. 
        The best-performing run (Run 66, sMAPE = 2.82\%) is highlighted in red.
        Each optimization run took an average of 1 minute and 35 seconds to complete.
        The panels show results for runs 1-25 (top-left), 26-50 (top-right), 51-75 (bottom-left), and 76-100 (bottom-right).
    }
    \label{fig:appendix-smape-grid} 
\end{figure}

In Figures~\ref{fig:vae-convergence}–\ref{fig:cvae-convergence}, we report the convergence of the VAE and CVAE as a function of optimization iterations and latent samples, respectively. We show in Figure~\ref{fig:vae-convergence} the best VAE\,+\,CMA-ES run (Run 66, see Figure~\ref{fig:appendix-smape-grid}). The best Mean Squared Error (MSE) found at each iteration of the optimizer drops rapidly during approximately the first 120 iterations, then enters a slower fine-tuning phase, and finally converges to a stable, accurate solution after approximately 260 iterations.

In Figure~\ref{fig:cvae-convergence}, we examine the convergence of the CVAE's functional sMAPE on a single test sample as the number of latent space samples increases. The cumulative average sMAPE stabilizes after approximately 10,000 samples, indicating that sampling beyond this point offers diminishing returns for estimating the mean prediction and its uncertainty. This result validates our use of 100,000 samples for the final CVAE evaluations in the main text, which ensures a stable and reliable assessment of the model's performance.

\clearpage
\section{Experimental Setup}
\label{appendix:setup}

\textbf{Hardware.}~All computational experiments, including dataset generation and model training, were conducted on a workstation equipped with two NVIDIA RTX 4060 Ti GPUs, each providing 16\,GB of VRAM and a theoretical peak performance of 22.06 TFLOPS (FP32). The system is powered by an AMD Ryzen 9 5950X 16-core processor and supported by 2 \(\times\) 32\,GB of DDR4 3200\,MHz DIMM RAM.

\textbf{Software.}~The project was developed on an Ubuntu 22.04.5 LTS system with package management handled by Conda. The computational workflow was distributed across distinct environments tailored for specific tasks, with primary library versions listed below:
\begin{itemize}
    \item \textbf{Dataset Generation:} The high-throughput data generation pipeline was executed in an environment running Python (v3.10.17) with JAX~\cite{jax2018github} (v0.6.2).
    \item \textbf{Model Training and Optimization:} The CVAE and MLP models were implemented and trained in an environment running Python (v3.12.7) and CUDA (v12.7). Key libraries included PyTorch~\cite{paszke2019pytorch} (v2.5.1), Optuna~\cite{Akiba2019Optuna} (v4.3.0) for hyperparameter optimization, and scikit-learn~\cite{scikit-learn} (v1.5.1) for baseline metrics.
    \item \textbf{Model Inference:} For inference and model analysis, the environment was updated with Python (v3.12.11), JAX~\cite{jax2018github} (v0.7.0), and PyTorch~\cite{paszke2019pytorch} (v2.8).
    \item \textbf{GPU-Accelerated Baseline:} The XGBoost baseline was trained in a dedicated RAPIDS environment (v25.08) running Python (v3.12.11). This provided GPU-accelerated versions of XGBoost (v3.0.3), cuDF (v25.08.00), and CuPy (v13.5.1) for end-to-end data handling on the GPU.
\end{itemize}
Across all environments, data manipulation relied on Pandas~\cite{mckinney-proc-scipy-2010}, and all figures were generated using Matplotlib~\cite{Hunter2007} and Seaborn~\cite{Waskom2021_seaborn}.

\clearpage
\section{Limitations}
\label{sec:limitations}

While this work introduces a powerful framework for the inverse design of frictional metainterfaces, it is important to acknowledge its limitations, which provide clear avenues for future research. We categorize these into three main areas: the underlying physical model, the generative framework itself, and the sim-to-real gap.

\subsection{Limitations of the Physical Model}
Our framework is built upon the Greenwood and Williamson (GW) contact mechanics model, which, while foundational, carries several simplifying assumptions.

\textbf{Material and Contact Assumptions.} The current model is parameterized for a specific Polydimethylsiloxane (PDMS)-on-glass interface under dry friction conditions. It does not account for adhesion, lubrication, viscoelasticity, or plasticity, which are critical in many real-world tribological systems.
    
\textbf{Statistical Representation.} The model treats the surface as a statistical distribution of hemispherical asperities. This idealization does not capture complex, non-spherical asperity shapes or the spatial arrangement of asperities, which can influence contact behavior.
    
\textbf{Static Conditions.} The forward model simulates forces under quasi-static conditions and does not capture dynamic effects like stick-slip phenomena or the velocity dependence of friction.

\subsection{Limitations of the Generative Framework}

\textbf{Need for Post-Processing.} The raw CVAE outputs are not guaranteed to be physically valid (e.g., mixture weights summing to one). While our post-processing step (Appendix~\ref{appendix:clamping}) ensures validity, it is an external correction. A more elegant solution would involve architectures that can respect these constraints intrinsically.
    
\subsection{Limitations of the Dataset and Sim-to-Real Gap}

The translation of in-silico designs to practical applications is the ultimate goal. Two primary challenges remain:

\textbf{Sim-to-Real Transfer.} While our initial zero-shot transfer to an experimental OOD target is promising, the framework must be rigorously validated through a campaign of fabrication and physical testing to quantify its real-world accuracy and robustness.
    
\textbf{Manufacturing Constraints.} Our design space exploration, while comprehensive, does not account for manufacturability. Physical fabrication processes have inherent limitations on the precision, range, and complexity of surface topographies they can produce. Integrating these constraints into the generative model or as a post-design filter is a critical next step.

\end{document}